\documentclass[11pt]{article}
\usepackage{appendix}
\usepackage{booktabs}
\usepackage{multirow}
\usepackage{makecell}
\usepackage{verbatimbox}
\usepackage{subfigure}
\usepackage{graphicx}
\usepackage{xspace}
\usepackage{bm}
\usepackage{bbm}
\usepackage{mathtools}
\usepackage{mathrsfs}
\usepackage{diagbox}
\usepackage{xcolor}
\usepackage{amsmath}
\usepackage{amsthm,amsmath,amssymb}
\usepackage{tabularx}
\usepackage{amsfonts}
\usepackage{mathrsfs}
\usepackage{colortbl}  
\usepackage{algorithmic}
\usepackage{algorithm}

\usepackage{svg}

\newcommand{\eg}{\emph{e.g.},\xspace}
\newcommand{\ie}{\emph{i.e.},\xspace}
\newcommand{\etc}{etc.\xspace}

\newcommand\figref[1]{Fig.~\ref{#1}}

\newcommand\tabref[1]{Table~\ref{#1}}

\newcommand\secref[1]{Sec.~\ref{#1}}

\newcommand{\fakeparagraph}[1]{\vspace{1mm}\noindent\textbf{#1.}}

\newcommand{\sysname}{\textsc{CoCo}\xspace }

\usepackage{EMNLP2023}
\usepackage{times}
\usepackage{latexsym}
\usepackage{svg} 

\usepackage[T1]{fontenc}

\usepackage[utf8]{inputenc}

\usepackage{microtype}

\usepackage{inconsolata}

\title{\sysname: Coherence-Enhanced Machine-Generated Text Detection Under Low Resource With Contrastive Learning}

\author{Xiaoming Liu\textsuperscript{$1,\dagger,\ast$}, Zhaohan Zhang\textsuperscript{$1,2,\dagger$}, Yichen Wang\textsuperscript{$1,\dagger$}, Hang Pu\textsuperscript{$1$}, Yu Lan\textsuperscript{$1$},  Chao Shen\textsuperscript{$1$} \\
        \textsuperscript{1}Faculty of Electronic and Information Engineering, Xi'an Jiaotong University \\
        No.28, Xianning West Road, Xi'an, China\\
        \textsuperscript{2}Queen Mary University of London, London, UK \\
        \texttt{
        \{xm.liu,ylan2020,chaoshen\}@xjtu.edu.cn, 
        \{zzh1103,yichen.wang,hpu2022\}@stu.xjtu.edu.cn}\\
        \textsuperscript{$\dagger$} Equal contribution, \textsuperscript{$\ast$} Corresponding author
        }
  
\begin{document}
\maketitle
\begin{abstract}
Machine-Generated Text (MGT) detection, a task that discriminates MGT from Human-Written Text (HWT), plays a crucial role in preventing misuse of text generative models, which excel in mimicking human writing style recently.
The latest proposed detectors usually take coarse text sequences as input and fine-tune pre-trained models with standard cross-entropy loss.
However, these methods fail to consider the linguistic structure of texts.
Moreover, they lack the ability to handle the low-resource problem, which could often happen in practice considering the enormous amount of textual data online.
In this paper, we present a \textbf{co}herence-based \textbf{co}ntrastive learning model named \sysname to detect the possible MGT under the low-resource scenario.
To exploit the linguistic feature, we encode coherence information in the form of graph into the text representation.
To tackle the challenges of low data resources, we employ a contrastive learning framework and propose an improved contrastive loss for preventing performance degradation brought by simple samples.
The experiment results on two public datasets and two self-constructed datasets prove our approach outperforms the state-of-the-art methods significantly.
Also, we surprisingly find that MGTs originated from up-to-date language models could be easier to detect than these from previous models, in our experiments.
And we propose some preliminary explanations for this counter-intuitive phenomena.
All the codes and datasets are open-sourced.\footnote{Codes are available at \href{https://github.com/YichenZW/Coh-MGT-Detection}{https://github.com/YichenZW/Coh-MGT-Detection} and datasets are at \href{https://huggingface.co/datasets/ZachW/MGTDetect_CoCo}{https://huggingface.co/datasets/ZachW/MGTDetect\_CoCo}.}
\end{abstract}

\section{Introduction}


Thriving progress in the field of text generative models (TGMs) \cite{yang2019xlnet, kenton2019bert, liu2019roberta, keskar2019ctrl, lewis2020bart, brown2020language, gao2021making, Madotto2021FewShotBP, ouyang2022training, touvron2023llama, anil2023palm}, \eg ChatGPT\footnote{https://chat.openai.com} and GPT-4 \cite{OpenAI2023GPT4TR}, enables everyone to produce MGTs massively and rapidly.  
However, the accessibility to high-quality TGMs is prone to cause misuses, such as fake news generation \cite{zellers2019defending, yanagi2020fake, Huang2022FakingFN}, product review forging \cite{adelani2020generating}, and spamming \cite{tan2012spammer}, \etc
MGTs are hard to distinguish by an untrained human for their human-like writing style \cite{ippolito2020automatic} and the excessive amount \cite{grinberg2019fake}, which calls for the study of reliable automatic MGT detectors.

\begin{figure}[t]
  \centering
  \includegraphics[width=\linewidth]{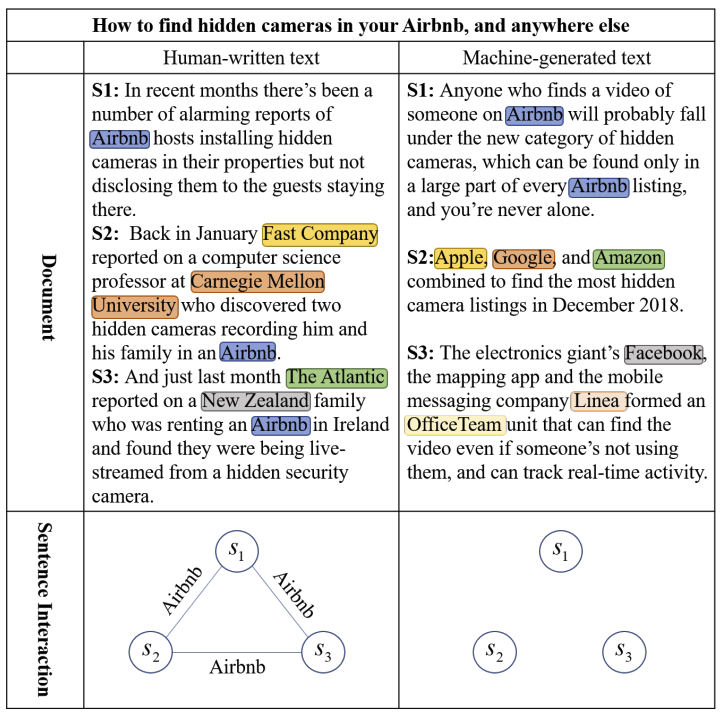}
  \caption{Illustration of sentence-level structure difference between HWT and MGT, the MGT is generated by GROVER \cite{zellers2019defending}.
  HWT is more coherent than MGT as the sentences share more same entities with each other.}
  \label{fig:coherent}
\end{figure}

Previous works on MGTs detection mainly concentrate on sequence feature representation and classification \cite{gehrmann2019gltr,solaiman2019release,zellers2019defending, he2023mgtbench, mitchell2023detectgpt}.
Recent studies have shown the good performance of automated detectors in a fine-tuning fashion \cite{solaiman2019release, mireshghallah2023smaller}.
Although these fine-tuning-based detectors have demonstrated their effectiveness, they still suffer from two issues that limit their conversion to practical use:
(1) 
Existing detectors treat input documents as flat sequences of tokens and use neural encoders or statistical features (\eg TF-IDF, perplexity) to represent text as the dense vector for classification. 
These fine-tuning-based methods rely much on the token-level distribution difference of texts in each class, which ignores high-level linguistic representation of text structure. 
(2)
Compared with the enormous number of online texts, the annotated dataset for training MGT detectors is rather low-resource.
Constrained by the amount of available annotated data, traditional detectors sustain frustrating accuracy and even collapse during the test stage.

 The defect in the coherence of LMs in generating long text has been revealed by previous works. 
\citet{malkin-etal-2022-coherence} mentions that long-range semantic coherence remains challenging in language generation. \citet{sun2020improving} also provides examples of incoherent MGTs. 
As shown in  \figref{fig:coherent}, MGTs and HWTs exhibit differences in terms of coherence traced by entity consistency. 
Accordingly, we propose that coherence could be an entry point for MGT detection via the perspective of high-level linguistic structure representation, where MGTs could be less interactive than HWTs.
Specifically, we propose an entity coherence graph to model the sentence-level structure of texts based on the thoughts of  Centering Theory \cite{grosz1986attention}, which evaluates text coherence by entity consistency.
The entity coherence graph treats entities as nodes and builds edges between entities in the same sentences and the same entities among different sentences to reveal the text structure.
Instead of treating text as a flat sequence, coherence modeling helps to introduce distinguishable linguistic features at the input stage and provides explainable differences between MGTs and HWTs.

To alleviate the low-resource problem in the second issue, inspired by the resurgence of contrastive learning \cite{he2020momentum,chen2020improved}, we utilize the proper design of sample pair and contrastive process to learn fine-grained instance-level features under low resource.
However, it has been proven that the easiest negative samples are unnecessary and insufficient for model training in contrastive learning \cite{cai2020all}.
To circumvent the performance degradation brought by the easy samples, we propose a novel contrastive loss with the capability to reweight the effect of negative samples by difficulty score to help the model concentrate more on hard samples and ignore the easy samples.
Extensive experiments on multiple datasets (GROVER, GPT-2, GPT-3.5)  demonstrate the effectiveness and robustness of our proposed method.
Surpirsingly, we find that the GPT-3.5 datasets are easier for all the detectors compared with datasets of smaller and older models (GPT-2 and GROVER) under our setting. We take a small step to exploring why the GPT-3.5 dataset is overly simple by probing statistical cues, including perspective from token spans and individual tokens.

In summary, our contributions are summarized as follows:

\begin{itemize}

\item \textbf{Coherence Graph Construction:}
We model the text coherence with entity consistency and sentence interaction while statistically proving its distinctiveness in MGT detection, and we further introduce the linguistic feature at the input stage.

\item \textbf{Improved Contrastive Loss:}
We propose a novel contrastive loss in which hard negative samples are paid more attention to improve the detection accuracy of challenging samples.

\item \textbf{Outstanding Performance:} 
We achieve state-of-the-art performance on four MGT datasets in both low-resource and high-resource settings. 
Experimental results verify the effectiveness and robustness of our model.

\end{itemize}
\section{Related Work}

\begin{figure*}[t]
  \centering
  \makebox[\textwidth][c]{\includegraphics[width=0.9\linewidth]{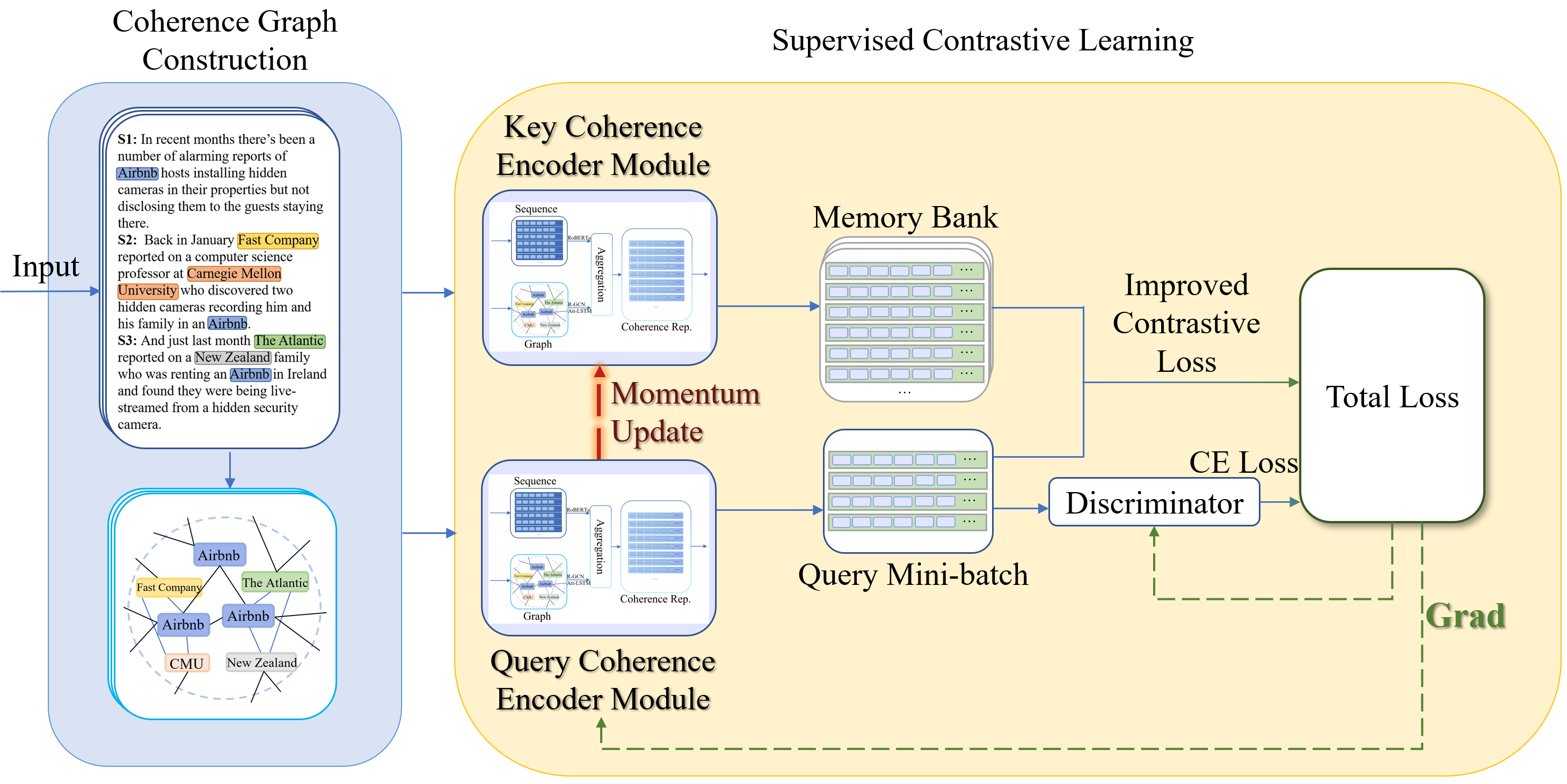}}
  \caption{ Overview of \sysname. Input document is parsed to construct a coherence graph (\ref{sec:1}), the text and graph are utilized by a supervised contrastive learning framework (\ref{sec:supcon}), in which coherence encoding module is designed to  encode and aggregate to generate coherence-enhanced representation (\ref{sec:2}). After that, we employ a MoCo based contrastive learning architecture in which key encodings are stored in a dynamic memory bank (\ref{dmb}) with improved contrastive loss to make final prediction (\ref{sec:3}). \\ }
  \label{fig:structure}
\end{figure*}

\fakeparagraph{Machine-generated Text Detection}
Machine-generated texts, also named deepfake or neural fake texts, are generated by language models to mimic human writing style, making them perplexing for humans to distinguish \cite{ippolito2020automatic}. 
Generative models like GROVER \cite{zellers2019defending}, GPT-2 \cite{radford2019language}, GPT-3 \cite{brown2020language}, and emerging GPT-3.5-turbo (also known as ChatGPT) have been evaluated on the MGT detection task and achieve good results \cite{gehrmann2019gltr, mireshghallah2023smaller}.
\citet{bakhtin2019real} train an energy-based model by treating the output of TGMs as negative samples to demonstrate the generalization ability.
Deep learning models incorporating stylometry and external knowledge are also feasible for improving the performance of MGT detectors \cite{uchendu2019characterizing, zhong2020neural}.
Our method differs from the previous work by analyzing and modeling text coherence as a distinguishable feature and emphasizing performance improvement under low-resource scenarios.

\fakeparagraph{Coherence Modeling}
For generative models, coherence is the critical requirement and vital target \citep{ hovy1988planning}. 
Previous works mainly discuss two types of coherence, local coherence \citep{ mellish1998experiments, althaus2004computing} and global coherence \citep{ mann1987rhetorical}. 
Local coherence focus on sentence-to-sentence transitions \citep{ lapata2003probabilistic}, while global coherence tries to capture comprehensive structure \citep{ karamanis2002stochastic}.
Our method strives to represent both local and global coherence with inner- and inter-sentence relations between entity nodes.

\fakeparagraph{Contrastive Learning}
Contrastive learning in NLP demonstrates superb performance in learning token-level embeddings \cite{su2022tacl} and sentence-level embeddings \cite{gao2021simcse} for natural language understanding.
With an in-depth study of the mechanism of contrastive learning, the hardness of samples is proved to be crucial in the training stage. 
\citet{cai2020all} define the dot product between the queries and the negatives in normalized embedding space as hardness and figured out the easiest 95\% negatives are insufficient and unnecessary.
\citet{song2022supervised} propose a difficulty measure function based on the distance between classes and apply curriculum learning to the sampling stage.
Differently, our method pays more attention to hard negative samples for improving the detection accuracy of challenging samples.

\section{Methodology}

The workflow of \sysname mainly contains coherence graph construction and supervised contrastive learning discriminator. \figref{fig:structure} illustrates its overall architecture.
The pseudocode of the training process is shown in Algorithm \ref{alg:1}.

\begin{figure*}[t]
  \centering
  \includegraphics[width=0.9\linewidth]{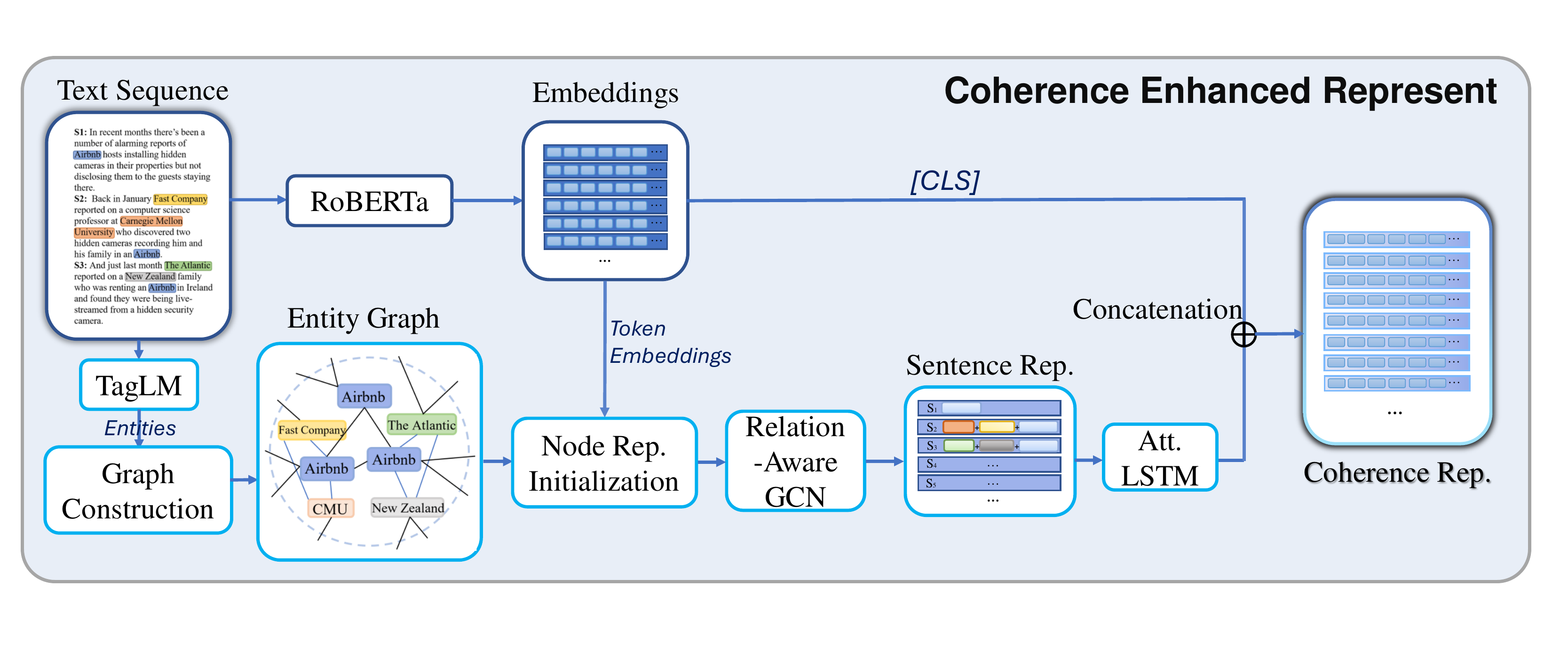}
  \caption{Illustration of CEM. It encodes and fuses the coherence graph and text sequence to generate coherence-enhanced representation of document.}
  \label{fig:encoder}
\end{figure*}

\subsection{Coherence Graph Construction}\label{sec:1}

In this part, we illustrate how to construct coherence graph to dig out the coherence structure of the text by modeling sentence interaction.

According to Centering Theory \cite{grosz1986attention}, the coherence of texts could be modeled by sentence interaction around center entities.
To better reflect text structure and avoid semantic overlap, we propose to construct an undirected graph with entities as nodes.
Specifically, we first implement the ELMo-based NER model TagLM \cite{peters2017semi} with the help of the NER toolkit AllenNLP\footnote{https://demo.allennlp.org/named-entity-recognition} to extract the entities from document.
A relation $<inter>$ is constructed between the same entities in different sentences and nodes within the same sentences are connected by relation $<inner>$ for their natural structure relevance.
Formally, the mathematical form of the coherence graph's adjacent matrix is defined as follows: 

\begin{small}
\begin{equation*}
\hspace{-2em}
     \bm{A}_{ij}=\left\{\begin{matrix} 1 & rel \  \mathtt{\langle inner \rangle}  & v_{i,a} \ne v_{j,b}, a=b  \\
                                 1 & rel \  \mathtt{\langle inter \rangle}  & v_{i,a} = v_{j,b} \ , a \ne b  \\
                                 0 & rel \ \ \textup{None}   & others \\ 
    \end{matrix}\right.
\end{equation*}
\end{small}

\noindent where $v_{i,a}$ represents $i$-th entity in sentence $a$, which is regarded as node in coherence graph.
We verify how MGT and HWT separate through static analysis on coherence graph in  Appendix \ref{app:static}.

\subsection{Supervised Contrastive Learning}\label{sec:supcon}
\subsubsection{Model Overview}
The training process is illustrated in \figref{fig:structure}.
Each entry in the dataset is documented with its coherence graph. The entries in the training set are sampled randomly into keys and queries.
Two coherence encoder modules (CEM) $f_k$ and $f_q$, are initialized the same to generate coherence-enhanced representation $\bm{D}_k$ and $\bm{D}_q$ for key and query.
A dynamic memory bank with the size of all training data is initialized to store all key representation and their annotations for providing enough contrastive pairs in low-resource scenarios.
In every training step, the newly encoded key graphs update the memory bank following the First In First Out (FIFO) rule to keep it updated and the training process consistent.
A novel loss composed of improved contrastive loss and cross-entropy loss ensures the model's ability to achieve instance-level intra-class compactness and inter-class separability while maintaining class-level distinguishability.
A linear discriminator takes query representations as input and generates prediction results.

\subsubsection{Positive/Negative Pair Definition}
In the supervised setting, where we have access to label information, we define two samples with the same label as positive pairs and those with different labels as negative pairs for incorporating label information into the training process.

\subsubsection{Encoder Design}\label{sec:2}
In this part, we introduce the structure of graph neural network structure, an innovative coherence encoder module(CEM), which is utilized to integrate coherence information into a semantic representation of text by propagating and aggregating information from different granularity.
The workflow is illustrated in \figref{fig:encoder}.

\fakeparagraph{Node Representation Initialization}
We initialize the representation of entity nodes with the powerful pre-trained model RoBERTa for its superior ability to encode contextual information into text representation. 

Given an entity $e$ with a span of $n$ tokens, we utilize RoBERTa to map input document $\bm{x}$ to embeddings $\bm{h}(\bm{x})$.
The contextual representation of $e$ is calculated as follows:
\begin{equation}
    \bm{Z}_{v}=\frac{1}{n}\sum_{i=0}^{n}\bm{h}(\bm{x})_{e_i},  
\end{equation}
where $e_i$ is the absolute position where the $i$-th token in $e$ lies in the whole document.

\fakeparagraph{Relation-aware GCN}
Based on the vanilla Graph Convolutional Networks \citep{welling2016semi}, we propose a novel method to assign different weight $\bm{W}_r$ for inter and inner relation $r$ with Relation-aware GCN.
Relation-aware GCN convolute edges of each kind of relation in the coherence graph separately. 
The final representation is the sum of GCN outputs from all relations. We use two-layer GCN in the model because more layers will cause an overfitting problem under low resources.
We define the relation set as $R$, and the calculation formula is as follows:

\begin{small}
\begin{equation}
\begin{split}
\bm{H}^{(i+1)} = \sum_{r\in R}^{} \hat{\bm{A}}&\text{ReLU}(( \hat{\bm{A}}\bm{H}^{(i)}\bm{W}_r^{(i)} )\bm{W}_r^{(i+1)}), \\
    \hat{\bm{A}} &= \tilde{\bm{D}}^{- \frac{1}{2}} \tilde{\bm{A}}\tilde{\bm{D}}^{-\frac{1}{2} },
\end{split}
\end{equation}
\end{small}
 where $\bm{H}^{(i)} \in \bm{R}^{N \times d}$ is node representation in $i$-th layer. $\tilde{\bm{A}} = \bm{A} + \bm{I}$, $\bm{A}$  is the adjacency matrix of the coherence graph, $\hat{\bm{A}}$ is the normalized Laplacian matrix of  $\tilde{\bm{A}}$, $\bm{W}_r$ is the relation transformation matrix for relation $r$.

\fakeparagraph{Sentence Representation}\label{fake:sent}
Afterward, we aggregate updated node representation from the last layer of Relation-aware GCN into sentence-level representation to prepare for concatenation with sequence representation from RoBERTa. 
The aggregation follows the below rule:

\begin{equation}
\bm{Z}_{s_i} = \frac{1}{M_i}\sum_{j}^{M_i}\sigma(\bm{W}_s\bm{H}_{(i,j)}+\bm{b}_s),
\end{equation}
where $M_i$ represents the number of entities in $i$-th sentence, $\bm{H}_{(i,j)}$ represents the embedding of $j$-th entity in $i$-th sentence, $\bm{W_s}$ is weight matrix and $\bm{b}_s$ is bias. All the sentence representations within the same document are concatenated as sentence matrix $\bm{Z}_s$. 

\fakeparagraph{Document Representation with Attention LSTM}
We design a self-attention mechanism for discovering the sentence-level coherence between one sentence and other sentences, and apply LSTM with the objective to track the coherence in continuous sentences and take the last hidden state of LSTM for aggregated document representation containing comprehensive coherence information. The calculation is described as follows:

\begin{small}
\begin{align}
    \bm{Z}_{c} = \text{LSTM}(\text{softmax}(\gamma \frac{\text{norm}(\bm{K})\text{norm}(\bm{Q})^T}{\sqrt{d_Z}})\bm{V}),
\end{align}
\end{small}

\noindent where $\bm{K}, \bm{Q}, \bm{V}$ are linear transformations of $\bm{Z}_s$ with matrix $\bm{W}_k, \bm{W}_q, \bm{W}_v$, $ d_Z $ is the dimension of representation $\bm{Z}_s$, and $ \gamma $ is a hypergammar-parameter for scaling. 

 Finally, we concatenate $\bm{Z}_{c}$ and the sequence representation $ \bm{h}(\text{[CLS]}) $ from the RoBERTa's last layer to generate document coherence-enhanced representation $\bm{D}$.

\subsubsection{Dynamic Memory Bank}
\label{dmb}
The dynamic memory bank is created to store as much as key encoding $\bm{D}_k$ to form adequate positive and negative pairs within a batch.
The dynamic memory bank is maintained as a queue so that the newly encoded keys can replace the outdated ones, which keeps the consistency between the key encoding and the current training step.

\subsubsection{Loss Function} \label{sec:3}
Following the definition of positive pairs and negative pairs above, traditional supervised contrastive loss \cite{gunelsupervised} treats all positive pairs and negative pairs equally.
However, with a recognition that not all negatives are created equal \cite{cai2020all}, our goal is to emphasize the informative samples to help the model differentiate difficult samples.
Thus, we propose an improved contrastive loss that dynamically adjusts the weight of negative pair similarity according to the hardness of negative samples.
To be specific, the hard negative samples should be assigned a larger weight to stimulate the model to pull the same classes together and push different classes away.
The improved contrastive loss is defined as:

\begin{small}
\begin{equation}
    \begin{split}
        \mathcal{L}_{\text{ICL}}=\sum_{j=1}^{M}\textbf{1}_{y_i=y_j}&\log\frac{S_{ij}}{\sum_{p\in \mathcal{P}(i)}^{}S_{ip}+\sum_{n\in \mathcal{N}(i)}^{}rf_{in}S_{in}   }, \\
        rf_{ij}&=\beta \frac{\bm{D}_{q}^{i}\bm{D}_{k}^{n}}{\text{avg}(\bm{D}_{q}^{i}\bm{D}_{k}^{1:|\mathcal{N}(i)|})}, \\
        S_{ij}&=\exp(\bm{D}_{q}^{i}\bm{D}_{k}^{j}/\tau),
    \end{split}
    \label{eq:icl}
\end{equation}

\end{small}
\noindent where $\mathcal{P}(i)$ is the positive set in which data has the same label with $q_i$ and $\mathcal{N}(i)$ is the negative set in which data has a different label from $q_i$.

Apart from instance-level learning mechanism, 
a linear classifier combined with cross-entropy loss $\mathcal{L}_{\text{CE}}$ is employed to provide the model with class-level separation ability.
$\mathcal{L}_{\text{CE}}$ is calculated by 
\begin{equation}
    \mathcal{L}_{\text{CE}}=\frac{1}{N} \sum\nolimits_{i=1}^{N}-[y_i\log(p_i)+(1-y_i)\log(1-p_i)],
    \label{eq:ce}
\end{equation}
where $p_i$ is the prediction probability distribution of $i$-th sample.  
The final loss $\mathcal{L}_{\text{total}}$ is a weighted average of $\mathcal{L}_{\text{ICL}}$ and $\mathcal{L}_{\text{CE}}$ as:
\begin{equation}
    \mathcal{L}_{\text{total}} = \alpha \mathcal{L}_{\text{ICL}} + (1-\alpha)\mathcal{L}_{\text{CE}},
    \label{eq:tot}
\end{equation}
where the hyperparameter $\alpha$ adjusts the relative balance between instance compactness and class separability.

\subsubsection{Momentum Update}
The parameters of query encoder $f_q$ and the classifier can be updated by gradient back-propagated from $\mathcal{L}_{\text{total}}$.
We denote the parameters of $f_q$ as $\theta_q$,
the parameters of $f_k$ as $\theta_k$,
The key encoder $f_k$'s parameters are updated by the momentum update mechanism:
\begin{equation}
    \theta_{k} \gets \beta \theta_k + (1-\beta)\theta_{q},
    \label{eq:mom}
\end{equation}
where the hyperparameter $\beta$ is momentum coefficient.

\begin{algorithm}
\small
	\caption{Algorithm of \sysname} 
	\label{alg:1}
	\begin{algorithmic}[1]
		\REQUIRE Input $X$, consisting of documents $D$ and corresponding coherence graph $G$, hyper-parameters such as the size of dynamic memory bank $M$ and batch size $S$, labels $Y$
		\ENSURE  A learned model \sysname, consisting of key encoder $f_k$ with parameters $\theta_k$, query encoder $f_q$ with parameters $\theta_q$, classifier $f_c$ with parameters $\theta_c$
		\STATE Initialize $\theta_k$ = $\theta_q$, $\theta_c$
		\STATE Initialize dynamic memory bank with $f_k(x_1, x_2...x_M)$, where $x_i$ is randomly sampled from $X$.
            \STATE Freeze $\theta_k$
		\STATE $epoch \leftarrow 0$
		\WHILE{$epoch\leq epoch_{\text{max}}$}
		\STATE $n \leftarrow 0$
		\WHILE{$n\leq n_{\text{max}}$}
		\STATE  Randomly select batch $\bm{b}_k$, $\bm{b}_q$
		\STATE $\bm{D}_q = f_q(\bm{b}_q)$, $\bm{D}_k = f_k(\bm{b}_k)$
            \STATE $\widehat{p} = f_c{(\bm{D}_q)}$
		\STATE Calculate $\mathcal{L}_{ICL}$ with equation \ref{eq:icl}, calculate $\mathcal{L}_{CE}$ with equation \ref{eq:ce}, calculate $\mathcal{L}_{total}$ with equation \ref{eq:tot}
		\STATE  Backward on $\mathcal{L}_{total}$ and update $\theta_q$, $\theta_c$ based on AdamW gradient descent with an adjustable learning rate
		\STATE Momentum update $\theta_k$ with equation \ref{eq:mom}
		\STATE Update dynamic memory bank $queue$ with $enqueue(queue, \bm{D}_k)$, $dequeue(queue)$
		\STATE $k \leftarrow k +1$
		\ENDWHILE
		\IF{ Early stopping }
		\STATE \textbf{break}
		\ELSE
		\STATE $epoch \leftarrow epoch +1$
		\ENDIF
		\ENDWHILE
		\RETURN  A trained model \sysname 
	\end{algorithmic}
\end{algorithm}
\section{Experiments}

\subsection{Datasets} \label{dataset}

\begin{table*}[t]
\renewcommand\arraystretch{1.5}
\centering
\resizebox{\textwidth}{!}{
\begin{tabular}{l|cccc|cccc}
\bottomrule
\multicolumn{1}{c|}{\textbf{Dataset}} & \multicolumn{4}{c|}{\textbf{GROVER}}                                                                                                              & \multicolumn{4}{c}{\textbf{GPT-2}}                                                                                            \\
\multicolumn{1}{c|}{\textbf{Size}}    & \multicolumn{2}{c}{Limited Dataset (500 examples)}                                                         & \multicolumn{2}{c|}{Full Dataset}                                         & \multicolumn{2}{c}{Limited Dataset (500 examples)}                                       & \multicolumn{2}{c}{Full Dataset}                                         \\ \cline{1-1}
\multicolumn{1}{c|}{\textbf{Metric}}   & ACC                                    & F1                                     & ACC                            & F1                            & ACC                           & F1                            & ACC                           & F1                            \\ \bottomrule
GPT2                                  & 0.5747 $ \pm $ 0.0217          & 0.4394 $ \pm $ 0.0346          & 0.8274  $ \pm $ 0.0091 & 0.8003 $ \pm $ 0.0141 & 0.5380 $ \pm $ 0.0067 & 0.4734 $ \pm $ 0.0182 & 0.8913 $ \pm $ 0.0066 & 0.8839 $ \pm $ 0.0078 \\
XLNet                                 & 0.5660 $ \pm $ 0.0265          & 0.4707 $ \pm $ 0.0402          & 0.8156 $ \pm $ 0.0079  & 0.7493 $ \pm $ 0.0073 & 0.6551 $ \pm $ 0.0083 & 0.5715 $ \pm $ 0.0095 & 0.9091 $ \pm $ 0.0091 & 0.9027 $ \pm $ 0.0111 \\
RoBERTa                               & 0.6621 $ \pm $ 0.0133          & 0.5895 $ \pm $ 0.0231          & 0.8772 $ \pm $ 0.0029  & 0.8171 $ \pm $ 0.0048 & 0.8223 $ \pm $ 0.0088 & 0.7978 $ \pm $ 0.0085 & 0.9402 $ \pm $ 0.0039 & 0.9384 $ \pm $ 0.0044 \\ \hline
DualCL             & \textit{0.5835 $ \pm $ 0.0857} & \textit{0.4628 $ \pm $ 0.1076} & \textit{0.7574 $ \pm $ 0.0855}                        & \textit{0.6388 $ \pm $ 0.1300}                              & \textit{0.6039 $ \pm $ 0.1367}             & \textit{0.5435 $ \pm $ 0.0903}                              & \textit{0.8023 $ \pm $ 0.1120}                             & \textit{0.8046 $ \pm $ 0.1530}                             \\ 
CE+SCL                                   & 0.6870 $ \pm $ 0.0142          & 0.5961 $ \pm $ 0.0197          & 0.8782 $ \pm $ 0.0044  & 0.8202 $ \pm $ 0.0057 & 0.8355 $ \pm $ 0.0046 & 0.8127 $ \pm $ 0.0067 & 0.9408 $ \pm $ 0.0006 & 0.9390 $ \pm $ 0.0009 \\ \hline
GLTR            & 0.3370 & 0.4935 & 0.6040 & 0.5182 & 0.7755 & 0.7639 & 0.7784 & 0.7691 \\
DetectGPT       & 0.5910 & 0.4258 & 0.6142 & 0.5018 & 0.7941 & 0.6982 & 0.7939 & 0.7002 \\ \hline
 \sysname      &  \textbf{ 0.6993  $ \pm $ 0.0119 }      &      \textbf{ 0.6125  $ \pm $ 0.0159 }                   &  \textbf{0.8826 $ \pm $ 0.0018}                          &  \textbf{0.8265 $ \pm $ 0.0036}                 &   \textbf{ 0.8530 $ \pm $ 0.0019}                          & \textbf{ 0.8410 $ \pm $ 0.0018}                        &    \textbf{ 0.9457 $ \pm $ 0.0004}                       &     \textbf{ 0.9452 $ \pm $ 0.0004}              \\ \bottomrule
 
\multicolumn{1}{c|}{\textbf{Dataset}} & \multicolumn{4}{c|}{\textbf{GPT-3.5 Unmixed}}                                                                                                              & \multicolumn{4}{c}{\textbf{GPT-3.5 Mixed}}                                                                                            \\
\multicolumn{1}{c|}{\textbf{Size}}    & \multicolumn{2}{c}{Limited Dataset (500 examples)}                                                         & \multicolumn{2}{c|}{Full Dataset}                                         & \multicolumn{2}{c}{Limited Dataset (500 examples)}                                       & \multicolumn{2}{c}{Full Dataset}                                         \\ \cline{1-1}
\multicolumn{1}{c|}{\textbf{Metric}}   & ACC                                    & F1                                     & ACC                            & F1                            & ACC                           & F1                            & ACC                           & F1                            \\ \bottomrule
GPT2              &               0.9023 $ \pm $ 0.0095 &  0.8920 $ \pm $ 0.0073    & 0.9917  $ \pm $ 0.0056 & 0.9905 $ \pm $ 0.0042 &  
0.8898 $ \pm $ 0.0094 &  0.8914 $ \pm $ 0.0084 &  0.9910 $ \pm $ 0.0046 & 0.9910 $ \pm $ 0.0033 \\
XLNet             &               0.9107 $ \pm $ 0.0068 &  0.9037 $ \pm $ 0.0064      & 0.9620  $ \pm $ 0.0043 & 0.9634 $ \pm $ 0.0068 &
0.8925 $ \pm $ 0.0106 &  0.8922 $ \pm $ 0.0089  & 0.9513 $ \pm $ 0.0052 & 0.9505 $ \pm $ 0.0039 \\
RoBERTa           &              0.9670 $ \pm $ 0.0084 &  0.9681 $ \pm $ 0.0077 & 0.9928  $ \pm $ 0.0035 & 0.9913 $ \pm $ 0.0040 &  
0.9565 $ \pm $ 0.0103 &  0.9583 $ \pm $ 0.0092  & 0.9923 $ \pm $ 0.0017 & 0.9901 $ \pm $ 0.0024 \\\hline
CE+SCL           &                0.9823 $ \pm $ 0.0053 &  0.9703 $ \pm $ 0.0070       & 0.9944  $ \pm $ 0.0023 & 0.9943 $ \pm $ 0.0031 & 
0.9628 $ \pm $ 0.0077 &  0.9686 $ \pm $ 0.0062 & 0.9932 $ \pm $ 0.0017 & 0.9905 $ \pm $ 0.0038 \\\hline

GLTR            & 0.9255 & 0.9287 & 0.9350 & 0.9358 & 0.9175 & 0.9181 & 0.9210 & 0.9212 \\

DetectGPT       & 0.9220 & 0.8744 & 0.9245 & 0.8991 & 0.8980 & 0.8814 & 0.9113 & 0.9041 \\ \hline

 \sysname  &  \textbf{ 0.9889 $ \pm $ 0.0044 }  &  \textbf{ 0.9791 $ \pm $ 0.0062 } &  \textbf{ 0.9972 $ \pm $ 0.0015 }  &  \textbf{ 0.9957 $ \pm $ 0.0020 }  &
 \textbf{ 0.9701 $ \pm $ 0.0069}   & \textbf{ 0.9735 $ \pm $ 0.0086 }  &  \textbf{ 0.9932 $ \pm $ 0.0019 }  &  \textbf{ 0.9937 $ \pm $ 0.0028 }              \\ \bottomrule
\end{tabular}
}
\caption{Results of the model comparison. It should be noticed that DualCL is easily affected by random seed, which may be caused by its weakness in understanding long texts. We do not present the experiment results for DualCL on GPT-3.5 dataset because the documents in GPT-3.5 dataset is so long that DualCL completely fails.}
\label{table:main_exp}
\vspace{-0.5cm}
\end{table*}

We evaluate our model on the following datasets:

     \textbf{GROVER Dataset} 
    \citep{zellers2019defending} is a News-style dataset in which HWTs are collected from RealNews, a large corpus of news from Common Crawl, and MGTs are generated by Grover-Mega (1.5B), a transformer-based news generator.
    
    \textbf{GPT-2 Dataset}
    is a Webtext-style dataset provided by OpenAI\footnote{https://github.com/openai/gpt-2-output-dataset} with HWTs adopted from WebText and MGTs produced by GPT-2 XLM-1542M.
    
     \textbf{GPT-3.5 Dataset} is a News-style open-source dataset constructed by us based on the text-davinci-003\footnote{https://platform.openai.com/docs/models/gpt-3-5} model (175B) of OpenAI, which is one of the most capable GPT-3.5 models so far and can generate longer texts (maximum 4,097 tokens). The GPT-3.5 model refers to various latest newspapers (Dec. 2022 - Feb. 2023) whose full texts act as the HWTs part, and the model generates by imitation. We design two subsets: \textbf{mixed-} and \textbf{unmixed-}provenances, whose details are explained in Appendix \ref{a1}. The brand-new datasets ensure no existing models have been pre-trained on the corpus, which accounts for the fairness of comparison.

The statistics of datasets are summarized in Appendix \ref{a0}.
We randomly sample 500 examples as training data for low-resource settings.
As for the full dataset setting, we utilize all training data.
The implementation details are in Appendix \ref{a3}. 

\subsection{Comparison Models}

We compare \sysname to state-of-the-art detection methods to reveal the effectiveness. 
We mainly divide comparison methods into two categories, \textbf{model-based} and \textbf{metric-based} methods. The metrics-based methods detect based on specific statistical text-evaluation metrics and logistic regression while the model-based methods learn features via fine-tuning a model.

The \textbf{model-based} baselines are as follows:

         \textbf{GPT-2} \citep{radford2019language}, \textbf{RoBERTa} \citep{liu2019roberta}, \textbf{XLNet} \citep{yang2019xlnet} are powerful transformers-based models fine-tuned on the binary classification task, implementing GPT-2 small(124M), RoBERTa-base(110M) and XLNet-base(110M).

       \textbf{CE+SCL} \cite{gunelsupervised},  a state-of-the-art supervised contrastive learning method in various downstream task. We train the detector with Cross-Entropy loss (CE) and supervised contrastive loss (SCL) calculated within a mini-batch.

       \textbf{DualCL} \cite{chen2022dual}, a contrastive learning method with the addition of label representations for data augmentation.

The \textbf{metric-based} baselines are as follows:

         \textbf{GLTR} 
        \cite{gehrmann2019gltr}, a supporting tool for facilitating humans to recognize MGTs with visual hints.
        We follow the settings of \cite{guo2023close} and select the Test-2 feature, which counts the top-$k$ tokens ranking from GPT-2 medium (355M) predicted probability distributions as features for training a logistic regression classifier.
        
         \textbf{DetectGPT} \cite{mitchell2023detectgpt}, a contemporaneous metric-based method utilizing the difference of model's log probability after text perturbations.
        We use T5-3B to perturb texts, and Pythia-12B \cite{biderman2023pythia} for scoring in the model. 
        A logistic regression classifier is trained to make predictions.

\subsection{Performance Comparison}

As shown in Table \ref{table:main_exp}, \sysname surpasses the state-of-the-art methods in MGT detection task by \textbf{at least} \textbf{1.23\%} and \textbf{1.64\%}, \textbf{1.75\%} and \textbf{2.83\%} on the GROVER, GPT-2 limited datasets in terms of Accuracy and F1-Score, respectively.
And \sysname achieves comparable performance with the most capable detectors in the complete dataset setting.
The result indicates the utility of contrastive learning and the rationality of coherence representation.

Moreover, it should be noticed that compared with metric-based methods, model-based methods usually tend to achieve better results.
This can be explained because metric-based methods can only concern and regress on a few features, which are over-compressed and under-represented for the detection task. 
Also, metric-based methods mainly use the pre-trained model for token probability instead of fine-tuning the whole model. 
And with more training samples involved, the performance of model-based methods improves drastically, while metric-based methods do not benefit much from more training examples.
It reveals that logistic regression is not strong enough to take in many texts with diverse semantics.
Meanwhile, \sysname outperforms CE+SCL and DualCL regardless of the size of the training set, which suggests the success of improved contrastive loss to solve the performance degradation problem brought by simple negative samples. 

We also find GROVER Dataset is the hardest to detect. 
It is because the GROVER generator is trained in an adversarial heuristic with the objective of deceiving the verifier, which endows the generator with a deceptive nature.
To our surprise, the GPT-3.5 dataset is overly simple for all detectors. The result is also in accord with conclusions in recent works \cite{mireshghallah2023smaller, chen2023gpt}.
We conduct extensive experiments on different self-constructed and published GPT-3.5 datasets generated by a series of prompts, validating this thundering conclusion.
The experiment details and results are in Appendix \ref{a2}.
We also implement experiments and discussions to explore further explanations in \secref{expl}.

Notably, a \textbf{more comprehensive comparison experiment} with 8 datasets \citep{pu2023zero} and 12 methods is presented in Appendix \ref{more_exp}, which substantiates the advantage of \sysname.

\subsection{Ablation Study\label{abl}}

To illustrate the necessity of components of \sysname, we conduct ablation experiments on 1,000-example GROVER dataset.
The ablation models' structure is as follows:

\begin{table}[!h]
\centering
\renewcommand\arraystretch{1.2}
\scalebox{0.97}{
\begin{tabular}{lcc}
\bottomrule
\textbf{Model}                           & \textbf{ACC}                 & \textbf{F1}                  \\ \bottomrule
\sysname{} \small{(Plain)}                          & \cellcolor[HTML]{77d0ff}0.7697                          & \cellcolor[HTML]{64baea}{0.6428}                        \\ 
\sysname{} \small{(Sentence Nodes)}                      & \cellcolor[HTML]{69c0f3}{0.7733}                        & \cellcolor[HTML]{6dc5fa}{0.6379}                        \\ 
\sysname{} \small{(Coherence)}                    & \cellcolor[HTML]{47a3d7}{0.7777}                        & \cellcolor[HTML]{47a3d7}{0.6463}                        \\ 
\sysname{} \small{(Coherence+LSTM)}                            & \cellcolor[HTML]{3d97cd}{0.7787}                        & \cellcolor[HTML]{3d97cd}{0.6471}                        \\ 
\sysname{} \small{(Coherence+LSTM+SCL)}                   & \cellcolor[HTML]{338bc3}{0.7827}                        & \cellcolor[HTML]{338bc3}{0.6609}                        \\ \hline
\textbf{\sysname}         & \cellcolor[HTML]{2980b9}{0.7843}                            & \cellcolor[HTML]{2980b9}{0.6684}                           \\ \bottomrule
\end{tabular}
}
\caption{Results of the ablation study on 1,000-example GROVER dataset.}
\label{table:ab}
\end{table}

\noindent \textbf{\sysname (Plain)} removes graph information and encodes only by RoBERTa parts. The model removes contrastive learning and only uses CE loss.

\noindent \textbf{\sysname (Sentence Nodes)} treats sentences (instead of entities) as nodes and establishes edges between sentences that share the same entities.
Node representation is initialized by RoBERTa embedding and mean-pooling operation.
Document representation is obtained by one CEM discarding sentence representation and attention LSTM part in \secref{fake:sent}.
Document representation is calculated by mean-pooling operation on sentence node representations.
A linear classification head with cross-entropy loss is used for detection.

\noindent \textbf{\sysname (Coherence)} incorporates the coherence graph into the representation of document and deploys the sentence representation of \secref{fake:sent}.
The rest are the same with \sysname (Sentence Nodes).

\noindent \textbf{\sysname (Coherence+LSTM)} uses attention LSTM for document-level aggregation, and the rest is the same as \sysname (Coherence).

\noindent \textbf{\sysname (Coherence+LSTM+SCL)} utilizes the contrastive learning framework, but the loss function is traditional supervised contrastive loss (SCL) instead of the improved contrastive loss.

As shown in \tabref{table:ab}, coherence information and the contrastive learning framework greatly contribute to the development of model performance, especially in F1-Score.
Replacing entity nodes in the coherence graph with sentences impairs the detector, which could be caused by semantic overlap between graph representation and text sequence representation.
The attention LSTM also plays an important role in preserving the coherence information during sentence aggregation.
Lastly, the results show the advantage of improved contrastive loss over standard supervised contrastive loss.

Furthermore, we also conduct ablation studies on other scenarios, including GPT-2, GPT-3.5-Unmixed, and GPT-3.5-Mixed	datasets. More detailed results are discussed in the Appendix \ref{app:abl}, which clearly stands for the performance gain of \sysname{} components. Moreover, the helpfulness of contrastive learning is verified to be orthogonal to the helpfulness of coherence information.	

\subsection{Discussion}

\subsubsection{Model Robustness to Perturbation}
To validate the robustness of \sysname to various perturbations, we train \sysname on the GROVER dataset in the low-resource setting and perturb the test set with four different operations: \textbf{Delete} (randomly delete tokens in each entry), \textbf{Repeat} (randomly select tokens and repeat them twice in the text), \textbf{Insert} (add random tokens from the vocabulary of the pre-trained model into random positions in the text), \textbf{Replace} (randomly replace tokens with randomly selected tokens from the vocabulary).
The perturbation scale is set to 15\%.
The experiment result is shown in Table \ref{table:robust}.



\begin{table}[h]
\renewcommand\arraystretch{1.5}
\centering
\scalebox{0.65}{
\begin{tabular}{ccccc}
\bottomrule
Model    & \multicolumn{2}{c}{RoBERTa}       & \multicolumn{2}{c}{\sysname} \\
Metric   & Acc             & F1              & Acc         & F1         \\ \hline
Original & 0.6635          & 0.5901          & \textbf{0.6993}           & \textbf{0.6125}          \\ \hline
\textbf{Delete}   & \thead{0.5736 (-0.0899)} & \thead{0.5545 (\textbf{-0.0356})} & \thead{\textbf{0.6363} (\textbf{-0.0630})}  & \thead{\textbf{0.5703} (-0.0422)}  \\
\textbf{Repeat}   & \thead{0.6320 (-0.0315)} & \thead{0.5743 (-0.0158)}  & \thead{\textbf{0.6732} (\textbf{-0.0261})}           & \thead{\textbf{0.6004} (\textbf{-0.0121})}          \\
\textbf{Insert}   & \thead{\textbf{0.6325} (\textbf{-0.0310})} & \thead{0.4881 (\textbf{-0.1020})} & \thead{0.6286 (-0.0707)}           & \thead{\textbf{0.4970} (-0.1155)}          \\
\textbf{Replace}  & \thead{0.5554 (-0.1081)} & \thead{0.4814 (\textbf{-0.1087})} & \thead{\textbf{0.6367} (\textbf{-0.0626})}           & \thead{\textbf{0.5023} (-0.1102)}          \\ \hline
Average  & \thead{0.5984 (-0.0651)} & \thead{0.5246 (\textbf{-0.0655})} & \thead{\textbf{0.6437} (\textbf{-0.0556})}           & \thead{\textbf{0.5425} (-0.0700)}         \\ \bottomrule
\end{tabular}
}
\caption{Model robustness to different perturbations.}
\label{table:robust}
\vspace{-0.5cm}
\end{table}

Despite the structural complexity, \sysname keeps outperforming the baseline during perturbations. \sysname's performance fluctuations are as minor as the baseline. And \sysname maintains \textbf{4.53\%} better in accuracy and  \textbf{1.79\%} better in F1-score on average, which stands for its robustness.

\subsubsection{Statistic Cues for Detectable Feature in GPT-3.5}
\label{expl}
To further investigate the rationale behind the easy-to-detect nature of GPT-3.5 generated texts, we utilize Transformers-Interpret\footnote{https://github.com/cdpierse/transformers-interpret}, a tool for evaluating feature attribution in predictions based on Integrated Gradients \cite{sundararajan2017axiomatic}, for discovering the supporters and opponents (tokens) in the decision-making stage.
We probe the statistical cues of the GPT-3.5 mixed dataset from two perspectives: spans of tokens and individual tokens.
We define spans of tokens coverage $\gamma_n$ as $n$-gram supporters for true positives $\mathbb{P}_n$, \ie  $n$ consecutive tokens all contribute positively to the correct prediction, over all n-gram tokens in true positives $\mathbb{A}_n$, which could be formulated as $\gamma_n=\frac{\mathbb{P}_n}{\mathbb{A}_n}$.

Moreover, we apply productivity $\pi_k$ and coverage $\epsilon_k$ of statistic cue $k$ \cite{niven2019probing} on the GPT-3.5 mixed dataset to find out if there are individual tokens acting as common and strong signals contribute to model predictions.
Formally, productivity $\pi_k$ is defined as:
\begin{equation}
    \begin{gathered}
       \alpha_k = \sum_{i=1}^{n}\mathbbm{1}[\exists j, k\in\mathbb{T}_j^{(i)}\wedge k\notin \mathbb{T}_{\neg j}^{(i)}] ,
     \\
    \pi_k = \frac{\sum_{i=1}^{n} \mathbbm{1}[\exists j, k\in \mathbb{T}_j^{(i)}\wedge k\notin\mathbb{T}_{\neg j}^{(i)}\wedge y_i=j] }{\alpha_k}.
    \end{gathered}
\end{equation}
Here, $\mathbb{T}_j^{(i)}$ is the set of tokens for text $i$ with label $j$.
And the coverage $\epsilon_k$ is the portion that all applicable cues over the total number of data points.

\begin{table}[]
\centering
\scalebox{0.85}{
\begin{tabularx}{\columnwidth}{c@{\hspace{1cm}}c@{\hspace{1cm}}c}
\bottomrule
  N-gram Coverage & MGT & HWT \\
  \hline      
$\gamma_1$ & 0.6659	& 0.6377 \\

$\gamma_2$ & 0.4250 & 0.3630 \\

$\gamma_3$ & 0.2883 & 0.2076 \\

$\gamma_4$ & 0.2019 & 0.1372 \\

$\gamma_5$ & 0.1425 & 0.0935 \\ 
\bottomrule
& &
\end{tabularx}
}
\vspace{-0.5cm}
\caption{N-gram Coverage in GPT-3.5 Mixed Dataset.}
\label{table:ngram coverage}
\end{table}

\begin{table}[]
\centering
\scalebox{0.9}{
\begin{tabularx}{\columnwidth}{c@{\hspace{1cm}}c@{\hspace{1cm}}c}
\bottomrule
Token     & Productivity & Coverage \\ \hline
\texttt{according} & 0.6923       & 0.3126   \\
\texttt{where}     & 0.6842       & 0.1998   \\
\texttt{they}      & 0.6316       & 0.3837   \\ \bottomrule
\end{tabularx}
}
\caption{Individual tokens with top-3 productivity.}
\label{table:top3}
\vspace{-0.5cm}
\end{table}

We fine-tune the RoBERTa-base model with a classification head on the GPT-3.5 mixed dataset and quantify how tokens in GPT-3.5 mixed test data affect the model predictions with the criteria mentioned above.
The results are shown in Table \ref{table:ngram coverage} and Table \ref{table:top3}.
It could be noticed that although $\gamma_1$ for MGT and HWT is about the same, the gap widens from $\gamma_2$ to $\gamma_5$, indicating that more consecutive spans of tokens act as an indicator for MGT than HWT.
Table \ref{table:top3} shows that "\texttt{according}", "\texttt{where}", and "\texttt{they}" are top-3 strongest tokens for detection. 
However, we could not reach any valid conclusions from their semantics. 
Meanwhile, these tokens only cover a small portion of the total number of data points (less than 0.4), leading to the weak strength of the signal they provide.
Therefore, we come up with a hypothesis that the easy-to-detect nature of GPT-3.5 does not originate from specific token but from certain language patterns (could be demonstrated by a span of tokens).
The reason might be that advanced LLMs fit extremely well to the corpus so that it generates more general expressions, which could be much easier to be expected by fine-tuned detectors.
A case study for token importance illustration is shown in Appendix \ref{case:token}.

Further, we discuss more topics in the Appendix, \eg the effect of hyper-parameters (\ref{hyper-para}), case study (\ref{case}), static geometric analysis on coherence graph (\ref{app:static}),  and exploration on imbalanced data (\ref{imbalance}).


\section{Conclusion}
In this paper, we propose \sysname, a coherence-enhanced contrastive learning model for MGT detection.
We construct a novel coherence graph from the document and implement a MoCo-based contrastive learning framework to improve model performance in low-resource settings.
An innovative encoder composed of relation-aware GCN and attention LSTM is designed to learn the coherence representation from the coherence graph, which is further incorporated with the sequence representation of the document.
To alleviate the effect of unnecessary easy samples, we propose an improved contrastive learning loss to force the model to pay more attention to hard negative samples.
\sysname outperforms all detection tasks generated by GROVER, GPT-2, and GPT-3.5, respectively, in both low-resource and high-resource settings.
We also find the outputs from the advanced GPT-3.5 are more detectable and explore the rationale behind the phenomena through the perspective of spans of tokens and individual tokens.

\section*{Limitations}
In this work, we step forward to better distinguishing MGTs under the low-resource setting. 
However, several limitations still exist for the broader applications of this detector.
Firstly, MGTs are easier to generate and collect than HWTs, which may cause an imbalanced label distribution in the dataset.
And \sysname literally corrupts in extremely imbalanced data distribution condition, as shown in \ref{imbalance}.
Future work could build upon the contrastive learning method of \sysname with innovation on sampling strategy for harsh low-resource and imbalanced data settings.
Secondly, our method artificially generates a coherence graph for every entry, which is not efficient for larger datasets.
What's more, short text, codes, and mathematical proofs, which are hard to generate coherence graphs, are also limitedly detected by CoCo.
More distinctive and easy-to-calculate features are worth exploring for generating distinguishable representations for texts with efficiency while better understanding the essence of TGMs.
Thirdly, with instruct-based generation and human-in-loop fine-tuning models prevailing, the strategy and defect of TGMs change slightly but constantly. 
The entity relation with the same semantic granularity and concretization in this paper would not be enough to detect the high-quality content by TGMs in the future.
More generative and adaptive detection models should be considered.

\section*{Ethical Considerations}
We provide insight into the potential weakness of TGMs and publish the GPT-3.5 news datasets.
We understand that the discovery of our work can be viciously used to confront detectors.
And we understand that malicious users can copy the contents of our GPT-3.5 news dataset to disguise real news and publish them.
However, with the purpose of calling for attention to detecting and controlling possible misuse of TGMs, we believe our work will inspire the advancement of the stronger detector of MGTs and prevent all potential negative uses of language models.

Our work complies with the sharing \& publication policy of OpenAI\footnote{https://openai.com/api/policies/sharing-publication/} and all data we collect is in the public domain and licensed for research purposes.

\section*{Acknowledgements}
We thank all the reviewers, the area chair, Kevin Yang (UC Berkeley), and Prof. Pietro Liò (Univ. of Cambridge) for their helpful feedback, which aided us in greatly improving the paper.
This work is supported by National Natural Science Foundation of China (62272371, 62103323, U21B2018), Initiative Postdocs Supporting Program (BX20190275, BX20200270), China Postdoctoral Science Foundation (2019M663723, 2021M692565), Fundamental Research Funds for the Central Universities under grant (xhj032021013), and Shaanxi Province Key Industry Innovation Program (2021ZDLGY01-02).

\bibliography{anthology,custom}
\bibliographystyle{acl_natbib}

\appendix

\section{Basic Statistics of Datasets \label{a0}}

\begin{table}[h]
\renewcommand\arraystretch{1.15}
\centering
\scalebox{1.0}{
\begin{tabular}{cllll}
\toprule
\textbf{Dataset} & \textbf{Class}  & Train  & Valid & Test  \\ \midrule
\multirow{2}{*}{GROVER} & HWT & 5,000  & 2,000 & 8,000 \\
                        & MGT & 5,000  & 1,000 & 4,000 \\
\multirow{2}{*}{GPT-2}  & HWT & 25,000 & 5,000 & 5,000 \\
                        & MGT & 25,000 & 5,000 & 5,000 \\ \midrule
\multirow{2}{*}{\makecell{GPT-3.5\\Unmixed}} & HWT & 3,454 & 1,000 & 1,000 \\
                        & MGT & 3,454 & 1,000 & 1,000 \\  
\multirow{2}{*}{\makecell{GPT-3.5\\Mixed}}  & HWT & 3,032 & 1,000 & 1,000 \\
                        & MGT & 3,032 & 1,000 & 1,000 \\ \bottomrule
\end{tabular}
}
\caption{Basic statistics of datasets.}
\label{table:1}
\end{table}

\section{Details of GPT-3.5 Dataset \label{a1}}

GPT-3.5 Dataset for \sysname is our latest dataset for the MGT detection task. There are two subsets in the self-made dataset for easy analysis of the impact of provenance and writing styles: unmixed- and mixed provinces.
We use the text-davinci-003 model of OpenAI to generate MGT examples.
The maximum length of HWTs is 1,024 tokens, and the target generation length is set as 1,024 tokens.
Here is an example of the MGT data.
\begin{center}
\fcolorbox{black}{gray!10}{\parbox{0.95\linewidth}{
\footnotesize
\texttt{"title": "On Eve of World Cup, FIFA Chief Says, ‘Don’t Criticize Qatar; Criticize Me.’",}

\texttt{"text": "DOHA, Qatar. The president of world soccer's governing body on Saturday sought to blunt mounting concerns about the World Cup in Qatar with a strident defense of both the host country's reputation and FIFA's authority over its showpiece championship. ...... Citing statistics, history and even childhood to bolster his case, he at one point likened his own experience as a redheaded child of immigrants to Switzerland to the assimilation problems of gays in the Middle East, and defended the laws, customs and honor of the host country.",}

 \texttt{"authors": ["Tariq Panja"],}
 
 \texttt{"publish\_date": "2022-11-19 00:00:00",}
 
 \texttt{"source": "The New York Times",}
 
 \texttt{"url": "https://www.nytimes.com/2022/11/19/sports/ soccer/world-cup-gianni-infantino-fifa.html"}
}}
\end{center}

And the following data shows the corresponding MGT in the dataset.

\begin{center}
\fcolorbox{black}{gray!10}{\parbox{.95\linewidth}{
\footnotesize
\texttt{"title": "On Eve of World Cup, FIFA Chief Says, ‘Don’t Criticize Qatar; Criticize Me.’",}

\texttt{"text": "The 2022 FIFA World Cup in Qatar is fast approaching, and its organizing committee’s president, Gianni Infantino, is speaking out about the lingering criticism of the country hosting the event. ...... he said. “It is a once-in-a-lifetime opportunity for the region to show the world its values and aspirations, and it is vital that this event is seen as a celebration of football and a celebration of the region.”",}

 \texttt{"authors": "machine",}
 
 \texttt{"source": "The New York Times",}
 
 \texttt{"matched\_hwt\_id": 202,}
 
 \texttt{"label": "machine""}

}}
\end{center}

\subsection{Human Written Texts}
\fakeparagraph{Unmixed Subset}
The HWTs of the unmixed subset are all from The New York Times\footnote{https://www.nytimes.com/} to exclude the impact of writing style. The time span of our data is Nov 1, 2022 - Dec 25, 2022, making sure that no pre-trained model has learned them. We develop the crawler based on news-crawler\footnote{https://github.com/LuChang-CS/news-crawler}.

\fakeparagraph{Mixed Subset}
The HWTs of the mixed subset come from various sources, listed as \tabref{tb:data_source}.
The time span of the data is Jan 1, 2022 - Jan 7, 2023. We develop the crawler based on Newspaper3k\footnote{https://github.com/codelucas/newspaper}.

\begin{table}[]
\scalebox{0.9}{
\begin{tabular}{ll}
\bottomrule
\textbf{Name}   & \textbf{Website}                    \\ \hline
Kotaku          & https://kotaku.com                  \\
The Daily World & https://www.thedailyworld.com       \\
CNN             & https://edition.cnn.com             \\
BBC             & https://www.bbc.com                 \\
NBC News        & https://www.nbcnews.com             \\
Reuters         & https://www.reuters.com             \\
Huffpost        & https://www.huffpost.com            \\
Pando           & http://pandodaily.com               \\
Yahoo           & https://news.yahoo.com              \\
Sun Times       & https://chicago.suntimes.com/news   \\
Sfgate          & https://www.sfgate.com              \\
New Republic    & https://newrepublic.com             \\
Time            & https://time.com                    \\
Pcmag           & http://www.pcmag.com                \\
CNBC            & https://www.cnbc.com/world/         \\
News            & https://www.news.com.au/            \\
The Atlantic    & https://www.theatlantic.com/latest/ \\
\bottomrule
\end{tabular}
}
\caption{Data sources for the mixed subset.}
\label{tb:data_source}
\end{table}

The dataset is specifically designed for MGTs detection and improving generation models. 
The contents of dataset are obtained from official news websites and the names of indicidual people are not mentioned maliciously.
And we strongly reject using our dataset to create offensive content or peek at private information. 

\subsection{Machine Generated Texts}

As the GPT-3.5 and ChatGPT model need prompts to generate, we write hints for the generation models to generate texts that meet our news-style long text generation. The hints format is as follows, and the content is related to HWTs.

\begin{center}
\fcolorbox{black}{gray!10}{\parbox{.95\linewidth}{
\footnotesize
\texttt{Write a news more than 1000 words.\\The news is written by \{Authors\} from \{Source\} in \{date\}. Title is \{title\}.}
}}
\end{center}

\section{GPT-3.5 Dataset Generated by Different Prompts and Experiment Results}
\label{a2}

To further validate the conclusion that GPT-3.5 generated texts are easier to detect, we utilize CNN news as a reference and design different prompts for GPT-3.5 generation.
The principle is to provide as much information as possible to GPT-3.5 to alleviate the possible gap in semantics and in length.

\fakeparagraph{Keywords as Prompt (KP)}
We extract the keywords and entities with GPT-3.5-turbo and provide examples in original news to form the prompt for generation.
The prompt format is as follows.

Example prompt for generation.

\begin{center}
\fcolorbox{black}{gray!10}{\parbox{.95\linewidth}{
\footnotesize
\texttt{"role": "system", "content": "Extract all the keywords, entities, and examples in the following passage:"\\"role": "user", "content": \{text\}}
}}
\end{center}

Example prompt for generation.

\begin{center}
\vspace{0em}
\fcolorbox{black}{gray!10}{\parbox{.95\linewidth}{
\footnotesize
\texttt{
Generate a news passage.\\The news is written by \{Authors\} from \{Source\} in \{date\}.\\Title: Lionel Messi isn’t expected to be back with PSG until early January after World Cup success \\Keywords: exploring, mountains, space, Poorna Malavath, Kavya Manyapu, NASA, Mount Everest, Project Shakthi, girls' education, Ladakh, India, virgin peak, climbing, altitude sickness, safety, motivation, empowerment, education, gender gap, Mount Aconcagua, sponsorship.\\Entities: CNN, Poorna Malavath, Kavya Manyapu, NASA, Mount Everest, Project Shakthi, Ladakh, India, Mount Aconcagua, South America, World Bank.\\Examples: designing space suits, youngest ever woman to summit Mount Everest, climbed a 6,012m virgin peak, raise money to fund girls' education, difficulties of climbing a virgin peak, experiences of altitude sickness, purpose of Project Shakthi, India's Right to Education Act, sponsorship for underprivileged school children, scaling Mount Aconcagua, expanding sponsorship globally.\\The target length for generation is 731 tokens. Add as much details and examples as you can.\\News: }
}}
\end{center}

\fakeparagraph{Summary as Prompt (SP)}
We employ GPT-3.5-turbo to summarize the original texts. 
The compression ratio is set to $[0.3, 1.0]$, which means the summary is required to be longer than 0.3 of the length of original text and shorter than whole original text.
The generated summary is used as prompt and the format is as follows:

\begin{center}
\fcolorbox{black}{gray!10}{\parbox{.95\linewidth}{
\footnotesize
\texttt{
Generate a news based on the following abstract: \\Paris Saint-Germain's coach Christophe Galtier has stated that Lionel Messi is not expected to join the team until early January as he is spending time in Argentina following the World Cup. Kylian Mbappé, Neymar Jr. and Achraf Hakimi, who played for their respective national teams at Qatar 2022, could return to the team as long as they are physically and mentally fit... \\The news is written by Matias Grez from CNN in 2022-12-28 00:00:00. \\Title: Lionel Messi isn’t expected to be back with PSG until early January after World Cup success\\News: }
}}
\end{center}
\fakeparagraph{Outline as Prompt (OP)}
We also outline the skeleton of original texts by GPT-3.5-turbo and feed the outline into GPT-3.5 text-davinci-003.
The prompt format is as follows:

Prompt for extraction.
\begin{center}
\fcolorbox{black}{gray!10}{\parbox{.95\linewidth}{
\footnotesize
\texttt{"role": "system", "content": "Write a hierarchical multi-point outline for the paragraph."\\"role": "user", "content": \{text\}}
}}
\end{center}

Example prompt for generation.

\begin{center}
\fcolorbox{black}{gray!10}{\parbox{.95\linewidth}{
\footnotesize
\texttt{
News Title: There's a shortage of truckers, but TuSimple thinks it has a solution: no driver needed \\The news is written by Jacopo Prisco, CNN from CNN in 2021-07-15 02:46:59. \\Outline: \\I. TuSimple's plan for fully autonomous truck tests\\A. Reliability of software and hardware needs to improve\\B. Fully autonomous tests without human safety driver planned by end of year\\ C. Results will determine if company can launch trucks by 2024\\D. 7,000 trucks reserved in US alone\\II. TuSimple’s competition\\A. ... \\Add more details and examples.\\News:}
}}
\end{center}

We first remove the HWTs that do not have desired length (i.e., 200-1024 tokens). 
And we take half of the selected HWTs as references to formulate different prompts mentioned above and feed it into GPT-3.5 to get MGTs.
The MGTs are sampled by Gaussion Distribution of their lengths.
To avoid the possible label leakage brought by text length, we directly filter the no-reference HWTs according to the Gaussion Distribution of MGT lengths.

Besides the self-constructed datasets, we also utilize the published GPT-3.5 dataset TuringBench benchmark (abbraviate as GPT-3.5 (TB)) \cite{uchendu2020authorship} to validate the deceptiveness of GPT-3.5. The statistics of datasets we use is in Table \ref{table:8}.

\begin{table}[!ht]
\centering
\scalebox{0.75}{
\begin{tabular}{cl|llll}
\hline
\multicolumn{2}{c|}{Dataset}  & Train  & Valid & Test & \# of tokens\\ \hline
\multirow{2}{*}{GPT-3.5(KP)} & HWT & 446  & 148 & 148 & 427.96 $ \pm $ 45.49 \\
                        & MGT & 446  & 148 & 148 & 403.88 $ \pm $ 75.63 \\
\multirow{2}{*}{GPT-3.5(SP)}  & HWT & 446 & 148 & 148 & 427.96  $ \pm $ 45.49 \\
                        & MGT & 446 & 148 & 148 & 415.72  $ \pm $ 66.54\\ \hline
\multirow{2}{*}{GPT-3.5(OP)} & HWT & 446 & 148 & 148 & 427.96 $ \pm $ 45.49 \\
                        & MGT & 446 & 148 & 148 & 429.34 $ \pm $ 78.62 \\  
\multirow{2}{*}{GPT-3.5(TB)}  & HWT & 5,964 & 975 & 1915 & 236.17  $ \pm $ 72.96 \\
                        & MGT & 5,507 & 894 & 1763 & 147.29 $\pm$ 70.15\\ 
\hline
\end{tabular}
}
\caption{Statistics of GPT-3.5 datasets.}
\label{table:8}
\end{table}

\begin{table*}[]
\renewcommand\arraystretch{1.5}
\centering
\resizebox{\textwidth}{!}{
\begin{tabular}{c|cccccccc}
\bottomrule

\multicolumn{1}{c|}{Dataset}  & \multicolumn{2}{c}{\textbf{GPT-3.5 (KP)}} &  
\multicolumn{2}{c}{\textbf{GPT-3.5 (SP)}}&
\multicolumn{2}{c}{\textbf{GPT-3.5 (OP)}}&
\multicolumn{2}{c}{\textbf{GPT-3.5 (TB)}}
\\
\multicolumn{1}{c|}{\textbf{Metric}}   & ACC(val/test)  & F1(val/test)  & ACC (val/test)  & F1 (val/test)  & ACC (val/test)   & F1 (val/test) & ACC (val/test)  & F1 (val/test)                            \\ \bottomrule
GPT2                                  & 0.9914/0.9916         & 0.9916/0.9918          & 0.9890/0.9893   & 0.9885/0.9889  & 0.9925/0.9928 & 0.9923/0.9924 & 0.9884/0.5422* & 0.9880/0.6335* \\
RoBERTa                               &  0.9946/0.9950 & 0.9950/0.9952 &  0.9935/0.9941  & 0.9933/0.9937 & 0.9946/0.9943 & 0.9942/0.9940 & 0.9962/0.6406* & 0.9960/0.7273*  \\
\sysname                               & 0.9955/0.9950        & 0.9942/0.9945          & 0.9938/0.9941   &  0.9936/0.9940  & 0.9942/0.9943 & 0.9942/0.9943 & 0.9966* & 0.9970* \\ \bottomrule
\end{tabular}
}
\caption{Experiment of different detectors on different GPT-3.5 Dataset. * : The great performance difference between validation set and test set on GPT-3.5 (TB) are because the test set randomly sample 50\% of the words of each article in the dataset \cite{uchendu2021turingbench}. We do not test \sysname{} on GPT-3.5 (TB) for the reason that such operation greatly influences the coherence in texts. We provide an example of this in Table \ref{tb:com}.}
\label{table:9}
\end{table*}

We conduct experiments with 3 random seeds and the average results are shown in Table \ref{table:9}. 
Counterintuitively, even if we elaborate the prompts and eliminate the length difference between MGTs and HWTs, the detection results are still superior, even on outdated baselines like GPT-2. 
The conclusion might be counterintuitive, but texts generated by the most advanced and popular GPT-3.5 model are the easiest to detect.

\begin{table*}[]
\renewcommand\arraystretch{1.5}
\centering
\resizebox{\textwidth}{!}{

\begin{tabular}[\textwidth]{cc}
\bottomrule
\textbf{GPT-3.5 (TB)}   & \textbf{GPT-3.5 (OP)}                    \\ \hline
  \fcolorbox{black}{gray!10}{\parbox{.5\linewidth}{
\footnotesize
'.video : morne morkel press conference * cricbuzz.video : england cricbuzz.bevan leads scotland 's 21-man squad for their first ever test match against pakistan in edinburgh icc.chris rogers retires after champions trophy defeat : australian cricketer announces international retirement the sun.icc super eight teams : odi ranking results.bahrain host oman on sunday kitply hans vohra gold cup gulf today.icc results.new zealand series history : india v new zealandyazan mohsen qawasma : how bahrain caught

}}        &   \fcolorbox{black}{gray!10}{\parbox{.5\linewidth}{
\footnotesize

Recent changes to key international indexes have resulted in the unprecedented exclusion of Russian stocks at a “zero” price, causing further losses in Moscow’s already-dismal stock exchange. This exclusion has made Russia no longer an option for investors, prompting a shift to other emerging markets.\textbackslash{}n\textbackslash{}nThe dramatic shift was made in early March, when FTSE Russell and MSCI announced the removal of Russian stocks from their indexes due to the country’s escalating economic and geopolitical problems. Shortly after, the Moscow Exchange suspended trading, sending ripples through the market.\textbackslash{}n\textbackslash{}nThe possible default on Russian debt has Western investors further reconsidering their investments in Russia...
}}                \\
                                                      \\
\bottomrule
\end{tabular}
}
\caption{A comparison example between texts in test set of GPT-3.5 (TB) and GPT-3.5 (OP). The GPT-3.5 (TB) text shows great disorder while GPT-3.5 (OP) text is neat.}
\label{tb:com}
\vspace{-0.5cm}
\end{table*}


\section{Implementation Details}
\label{a3}
This part mentions the implementation details and hyper-parameter settings of all the methods in the experiment.
To imitate the situation of low data-resources, we randomly sample 500 entries  from the datasets as limited dataset (positive:negative=1:1), which will test models together with the complete datasets.
And we conduct experiments on 10 different seeds and report the average test accuracy, F1-Score, and standard deviation only for model-based methods because metric-based methods would not be affected by random seeds.

We use RoBERTa base model to initialize the embedding of our representation and optimize the model using AdamW \citep{loshchilov2018decoupled} optimizer with a 0.01 weight decay.
We set the initial learning rate to $10^{-5}$ and the batch size to $8$ for all datasets based on experiences.

We utilize packages, namely transformers, pytorch, and allennlp to implement \sysname.
And the GPT-3.5 datasets and ChatGPT case is generated by OpenAI API and websites. 
We spend \$300 for API costs, including development and final generation costs.
We train and do experiments on 8 NVIDIA A100 GPUs on 2 Ubuntu-based servers. The total budget for training 20 epochs, dev, and testing on the GROVER dataset is 2.5 hours. On GPT-2 dataset is 12 hours, and on GPT-3.5 dataset is 1.5 hours.
We will publish our code and dataset recently.

\begin{table*}[h]
\centering
\scalebox{0.86}{
\begin{tabular}{c l c c c c c c c c}
\toprule
 & \textbf{Dataset Generator} & \multicolumn{2}{c}{\textit{\textbf{GPT-2 md}}} & \multicolumn{2}{c}{\textit{\textbf{GPT-2 xl}}} & \multicolumn{2}{c}{\textit{\textbf{GPT-3}}} & \multicolumn{2}{c}{\textit{\textbf{GPT-4}}} \\
\textbf{Type} & \textbf{Method} & ACC & AUROC & ACC & AUROC & ACC & AUROC & ACC & AUROC \\
\midrule
\multirow{3}{*}{\makecell{probability\\metric-based}} & GLTR & 0.7840 & 0.8536 & 0.7360 & 0.8098 & 0.2780 & 0.1930 & 0.4320 & 0.3990 \\
 & Rank & 0.6680 & 0.7200 & 0.6160 & 0.6723 & 0.4520 & 0.4304 & 0.5120 & 0.5203 \\
 & LogRank & 0.8080 & 0.8837 & 0.7600 & 0.8374 & 0.2800 & 0.1988 & 0.4220 & 0.3885 \\
  \midrule
\multirow{2}{*}{\makecell{perturbed\\metric-based}} & DetectGPT-10d & 0.8620 & 0.8400 & 0.8020 & 0.8896 & 0.3100 & 0.2349 & 0.4020 & 0.3601 \\
 & DetectGPT-10z & 0.8480 & 0.8331 & 0.8200 & 0.8977 & 0.3120 & 0.2330 & 0.4020 & 0.3585 \\
 \midrule
\multirow{2}{*}{\makecell{off-the-shelf\\ model-based}} & OpenAI-detector & 0.8460 & 0.8341 & 0.7740 & 0.8680 & 0.4400 & 0.4263 & 0.4200 & 0.3936 \\
 & ChatGPT-detector & 0.4760 & 0.4957 & 0.4900 & 0.5156 & 0.9280 & 0.9764 & 0.8640 & 0.9013 \\
 \midrule
\multirow{6}{*}{\makecell{fine-tuned\\model-based}} & OpenAI-GPT & 0.8050 & 0.8278 & 0.8170 & 0.8189 & 0.8450 & 0.8460 & 0.9020 & 0.9026 \\
 & BERT-base & 0.8480 & 0.8480 & 0.8540 & 0.8543 & 0.8570 & 0.8599 & 0.9260 & 0.9275 \\
 & GPT-2 & 0.6680 & 0.7896 & 0.7300 & 0.7247 & 0.9920 & 0.9920 & 0.8990 & 0.9010 \\
 & RoBERTa-base & 0.8940 & 0.8941 & 0.8970 & 0.8978 & 0.9840 & 0.9840 & 0.9630 & 0.9630 \\
 & Electra-base & 0.8710 & 0.8726 & 0.8720 & 0.8727 & 0.8880 & 0.8940 & 0.9350 & 0.9351 \\
 & \textbf{CoCo} & \textbf{0.9067} & \textbf{0.9123} & \textbf{0.9063} & \textbf{0.9063} & \textbf{0.9936} & \textbf{0.9938} & \textbf{0.9787} & \textbf{0.9786} \\
 \toprule
 & \textbf{Dataset Generator} & \multicolumn{2}{c}{\textit{\textbf{GPT-Neo lg}}} & \multicolumn{2}{c}{\textit{\textbf{GPT-J}}} & \multicolumn{2}{c}{\textit{\textbf{LLaMA 7B}}} & \multicolumn{2}{c}{\textit{\textbf{LLaMA 13B}}} \\

\textbf{Type} & \textbf{Method} & ACC & AUROC & ACC & AUROC & ACC & AUROC & ACC & AUROC \\
\midrule
\multirow{3}{*}{\makecell{probability\\ metric-based}} & GLTR & 0.7240 & 0.8044 & 0.6940 & 0.7574 & 0.5980 & 0.6086 & 0.5840 & 0.6082 \\
 & Rank & 0.6660 & 0.7329 & 0.6420 & 0.6923 & 0.5760 & 0.6114 & 0.5660 & 0.6106 \\
 & LogRank & 0.7580 & 0.8449 & 0.7480 & 0.8300 & 0.6160 & 0.6465 & 0.6160 & 0.6468 \\
 \midrule
\multirow{2}{*}{\makecell{perturbed\\ metric-based}} & DetectGPT-10d & 0.6900 & 0.7545 & 0.7560 & 0.8271 & 0.5640 & 0.5877 & 0.5300 & 0.5481 \\
 & DetectGPT-10z & 0.6860 & 0.7483 & 0.7560 & 0.8434 & 0.5740 & 0.5931 & 0.5320 & 0.5570 \\
  \midrule
\multirow{2}{*}{\makecell{off-the-shelf\\ model-based}} & OpenAI-detector & 0.7620 & 0.8615 & 0.7200 & 0.7904 & 0.6140 & 0.6712 & 0.5940 & 0.6453 \\
 & ChatGPT-detector & 0.8400 & 0.8798 & 0.8500 & 0.8875 & 0.8440 & 0.8845 & 0.8480 & 0.8880 \\
 \midrule
\multirow{6}{*}{\makecell{fine-tuned\\model-based}} & OpenAI-GPT & 0.7480 & 0.7611 & 0.6720 & 0.6720 & 0.6100 & 0.6142 & 0.6330 & 0.6335 \\
 & BERT-base & 0.7390 & 0.7690 & 0.7200 & 0.7277 & 0.6460 & 0.6462 & 0.6430 & 0.6595 \\
 & GPT-2 & 0.8940 & 0.8954 & 0.8970 & 0.8990 & 0.7960 & 0.8046 & 0.9050 & 0.9100 \\
 & RoBERTa-base & 0.9270 & 0.9326 & 0.9220 & 0.9290 & 0.9180 & 0.9254 & 0.9240 & \textbf{0.9669} \\
 & Electra-base & 0.7880 & 0.8320 & 0.7740 & 0.7816 & 0.6690 & 0.6920 & 0.7060 & 0.7063 \\
 & \textbf{CoCo} & \textbf{0.9462} & \textbf{0.9353} & \textbf{0.9326} & \textbf{0.9414} & \textbf{0.9321} & \textbf{0.9313} & \textbf{0.9455} & 0.9602 \\
\bottomrule
\end{tabular}
}
\caption{Comprehensive experimental results on wide scenarios. The same as the limited setting in \secref{dataset}, which uses 500 examples for these models to fine-tune.}
\label{table:more_exp}
\vspace{-0.5cm}
\end{table*}

\section{More Comparison Experiments} \label{more_exp}
Provisioning empirical evidence to claim effectiveness is a relatively broad topic, and in Table \ref{table:main_exp} we have shown \sysname outperforms on 4 datasets (8 settings) compared with 6 models, including Roberta and CE+SCL, the SOTA of the model-based methods, and DetectGPT, the SOTA of the metric-based methods. Moreover, our model is outperforming on very wide scenarios. Due to the limitation of pages, we do not post all the results in the main text, so we would love to share with you a more comprehensive result here.

\fakeparagraph{Dataset} Following \citet{pu2023zero}, we use RealNews dataset \cite{raffel2020exploring} as human-written texts, and the machine-generators are the most representative models nowadays, namely GPT-2 (medium and xl) \cite{radford2019language}, GPT-3 (text-davinci-003) \cite{ouyang2022training}, GPT-4 \cite{OpenAI2023GPT4TR}, GPT-Neo (2.7B) \cite{black2022gpt}, GPT-J \cite{gpt-j}, and LLaMA (7B and 13B) \cite{touvron2023llama}.

\fakeparagraph{Comparison Models}
More detailed, current detection methods can be categorized into four types: probability metric-based, perturbed metric-based, off-the-shelf model-based, and fine-tuned model-based. Our model \sysname{} is in the fine-tuned model-based category.
\begin{itemize}
    \item \textbf{Probability metric-based methods:} GLTR \cite{gehrmann2019gltr}, \ie using token log-likelihood; Rank \cite{solaiman2019release} and LogRank \cite{ippolito2020automatic}, \ie using the rank/log-rank of token likelihood.
    \item \textbf{Perturbed metric-based methods:} DetectGPT \cite{mitchell2023detectgpt}, in the nomenclature of \tabref{table:more_exp}, the number ‘10’ means the number of perturbation samples. The letter ‘d’ means not normalized on distribution, while ‘z’ means normalized.
    \item \textbf{Off-the-shelf model-based model:} OpenAI-detector \cite{solaiman2019release}, built by OpenAI mainly for GPT-2 detection based on the RoBERTa model; 
ChatGPT-detector \cite{guo2023close}, made based on SimpleAI based on the HC3 dataset.
    \item \textbf{Fine-tuned model-based methods:} All the models we use have the same level of size, \ie around 110M parameters, including OpenAI-GPT, Bert-base-uncased, GPT-2, RoBERTa-base, Google Electra-base discriminator, and \sysname.
\end{itemize}

Table \ref{table:more_exp} reveals the outstanding performance of \sysname in almost all the scenarios. 
Moreover, We also find the following phenomenon:
\begin{itemize}
    \item It follows the intuitive notion that off-the-shelf models are only competitive in their designed scenario. OpenAI-detector performs well on GPT-2s and GPT-Neo datasets. And ChatGPT-detector, in reverse, excels on GPT-3, GPT-4, Llamas, GPT-J, and GPT-Neo.
    \item Probability metric-based methods rely on the likelihood from the generation model, which is mainly designed for white-box machine-generated detection. For white-box models like GPT-2, GPT-Neo, and GPT-J, their performance is relatively good. But when applied to totally black-box models, these methods could easily fail. 
    DetectGPT, the perturbed metric-based method, shares the same limitation with a similar mechanism. 
    \item Among all the fine-tuned model-based methods, RoBERTa-base shows the best performance average on all datasets compared to other base models. Thus, it supports our claim that recognizing RoBERTa as SOTA for this category, and further built CL methods and \sysname{} based on RoBERTa. 
\end{itemize}

\section{Effect of Hyper-Parameters\label{hyper-para}}
\label{a4}

\subsection{Contrastive Learning Parameters}

\begin{figure}[!h]
  \centering
  \includegraphics[width=\linewidth]{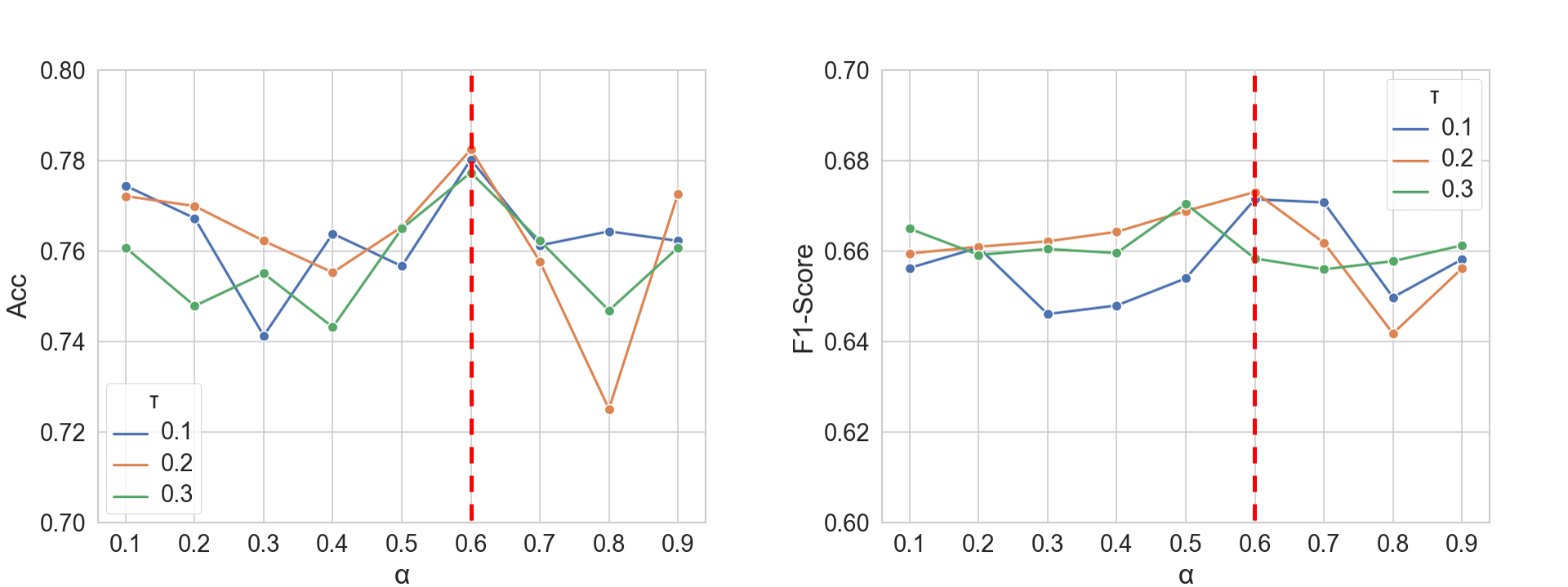}
  \caption{Effect of parameters $\alpha$ and $\tau$ on model performance.}
  \label{fig:modelpa}
  \vspace{-0.5cm}
\end{figure}

We evaluate the influence of contrastive learning hyper-parameters $\alpha$ and $\tau$ with experiments on different combinations of them. 
The result is shown in \figref{fig:modelpa}.
Considering the discovering that smaller $\tau$ leads to better hard negative mining ability \cite{wang2021understanding}, we select $\alpha$ from $\{0.1,0.2,...,0.9\}$ and $\tau$ from $\{0.1,0.2,0.3\}$.
We find that the extreme $\alpha$ value causes the performance degradation and the best hyper-parameter combination is $\alpha, \tau=0.6, 0.2$.
Our analysis is that large $\alpha$ forces the model to concentrate on the instance-level contrast and small $\alpha$ lets class separation objective take control.
Both will reduce the generalization performance of the detector on test set.

\subsection{Graph Parameters}

\begin{figure}[!h]
  \centering
  \includegraphics[width=1.1\linewidth]{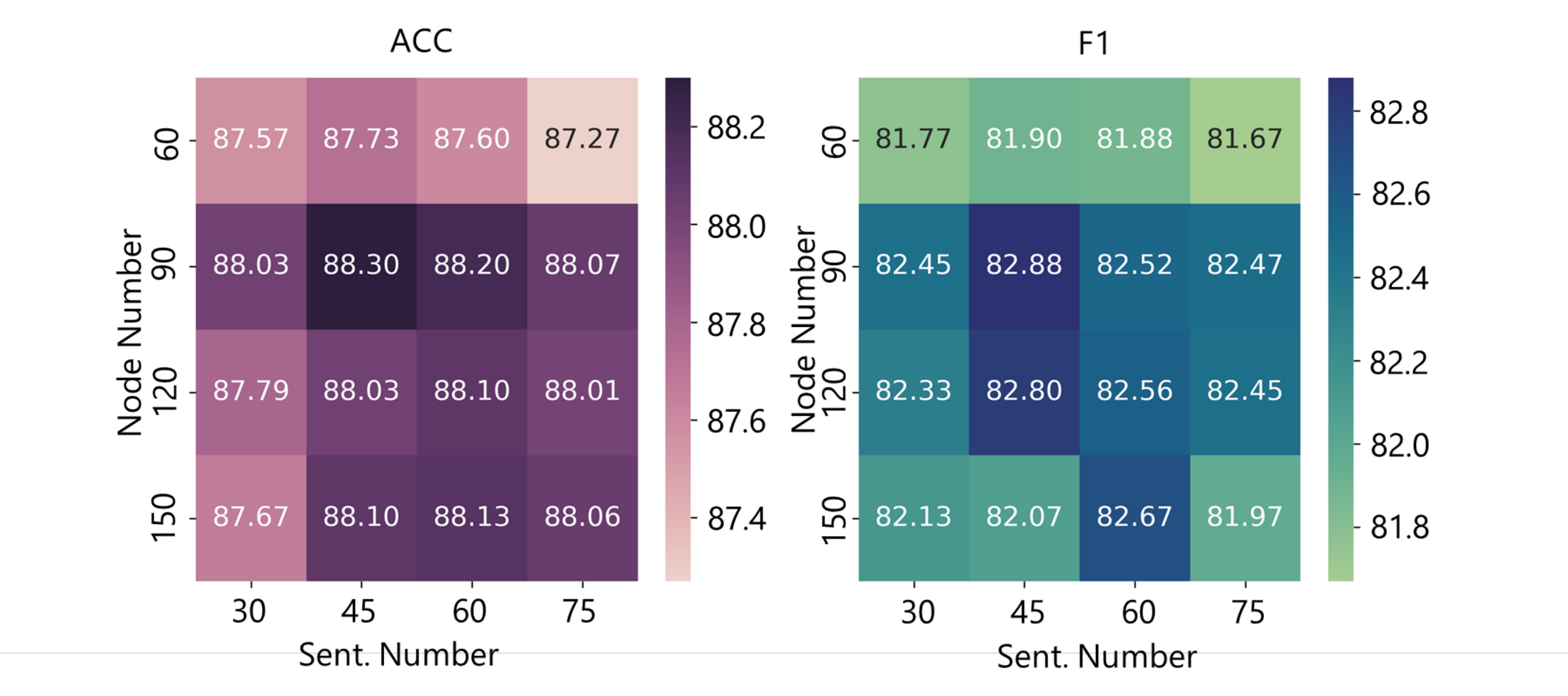}
  \caption{Performance of \sysname with different graph parameters.}
  \label{fig:graphpa}
  \vspace{-0.5cm}
\end{figure}

We further investigate the effect of max node number and max sentence number on model performance.
The result is shown in  \figref{fig:graphpa}.
We select max node number from $\{ 60,90,120,150 \}$ and max sentence number from $\{ 30,45,60,75 \}$.
The detector performs best when max node number is 90 and max sentence number is 45.
The experiment results prove that the large node and sentence number are not necessary for the improvement of detection accuracy.
We infer that even though setting large node and sentence number includes more entity information, excessive nodes bring noise to the model and impair the distinguishability of coherence feature.

\begin{table*}[h]
\centering
\renewcommand\arraystretch{1.2}
\scalebox{0.8}{
\begin{tabular}{lcccccccc}
\toprule
\textbf{Dataset}   & \multicolumn{2}{c}{\textbf{GPT-2}} & \multicolumn{2}{c}{\textbf{GPT-3.5 Unmixed}} & \multicolumn{2}{c}{\textbf{GPT-3.5 Mixed}} & \multicolumn{2}{c}{\textbf{Avg. Increase}} \\
\textbf{Metric}    & ACC  & F1     & ACC  & F1     & ACC  & F1    & ACC  & F1       \\ \midrule

\small{\sysname (Plain)}   & 0.8223      & 0.7978      &0.9670     &0.9681     &0.9565     &0.9583     &-      &-      \\ 
\small{\sysname (Coherence)}          & 0.8325      & 0.8217      & 0.9778   & 0.9785       & 0.9698        & 0.9704       & \cellcolor[HTML]{69c0f3}↑ 0.0114  & \cellcolor[HTML]{47a3d7}↑ 0.0154\\ 
\small{\sysname (Coherence + LSTM)}      & 0.8356        & 0.8274     &0.9778     &0.9787     &0.9703     &0.9710     & \cellcolor[HTML]{69c0f3}↑ 0.0126    &  \cellcolor[HTML]{338bc3}↑ 0.0176                  \\ 
\small{\sysname (ICL)}    & {0.8417}   & {0.8319}     &{0.9798}     &{0.9779}      &{0.9646}       &{0.9654}        &\cellcolor[HTML]{47a3d7}{↑ 0.0134}     &\cellcolor[HTML]{338bc3}{↑ 0.0170}                     \\ \hline
\textbf{\sysname}        & {0.8530}   & {0.8410}   &{0.9889}   &{0.9791}       &{0.9701}       &{0.9735}       &\cellcolor[HTML]{2980b9}{↑ 0.0220}    &\cellcolor[HTML]{2980b9}{↑ 0.0231}                 \\ \bottomrule
\end{tabular}
}
\caption{Results of ablation study on 500-example GPT-2, GPT-3.5-Unmixed, and GPT-3.5-Mixed datasets.}
\label{app:ab}
\vspace{-0.5cm}
\end{table*}

\section{Ablation Study\label{app:abl}}

In \secref{abl}, we mainly show the performance gain on the GROVER dataset. To further verify the effectiveness of \sysname{} across other scenarios. We also do the ablation study on 500-example GPT-2, GPT-3.5-Unmixed, and GPT-3.5-Mixed datasets. The result is shown in \tabref{app:ab}.	 

Here, we add a new ablated setting, \sysname{} (ICL), which applies the improved contrastive learning we proposed but does not include any part of the coherence graph representation model (\ie Coherence and LSTM). 

By comparing \sysname{} (Coherence) with \sysname{} (Plain), we can evaluate the effectiveness of the coherence model. It shows an average improvement of 1.14\% accuracy and 1.54\% F1 on the plain version. Furthermore, if we add attention LSTM for concatenation, it can achieve 1.26\% accuracy enhancement.

Moreover, by comparing \sysname{} (ICL) and \sysname{}, we further show the effectiveness of the coherence model based on the ICL model. There’s a gap of  0.86\% accuracy and 0.61\% F1 between with Coherence model and w/o it. The result shows the effectiveness of the coherence model component doesn’t heavily overlap with the effectiveness of the ICL method component. In conclusion, both two components of \sysname{} function effectively and cooperate with the other beneficially.

\section{Case Study\label{case}}

\subsection{Coherence Graph Difference\label{case:coherence}}
\begin{figure}[!h]
  \centering
  \includegraphics[width=\linewidth]{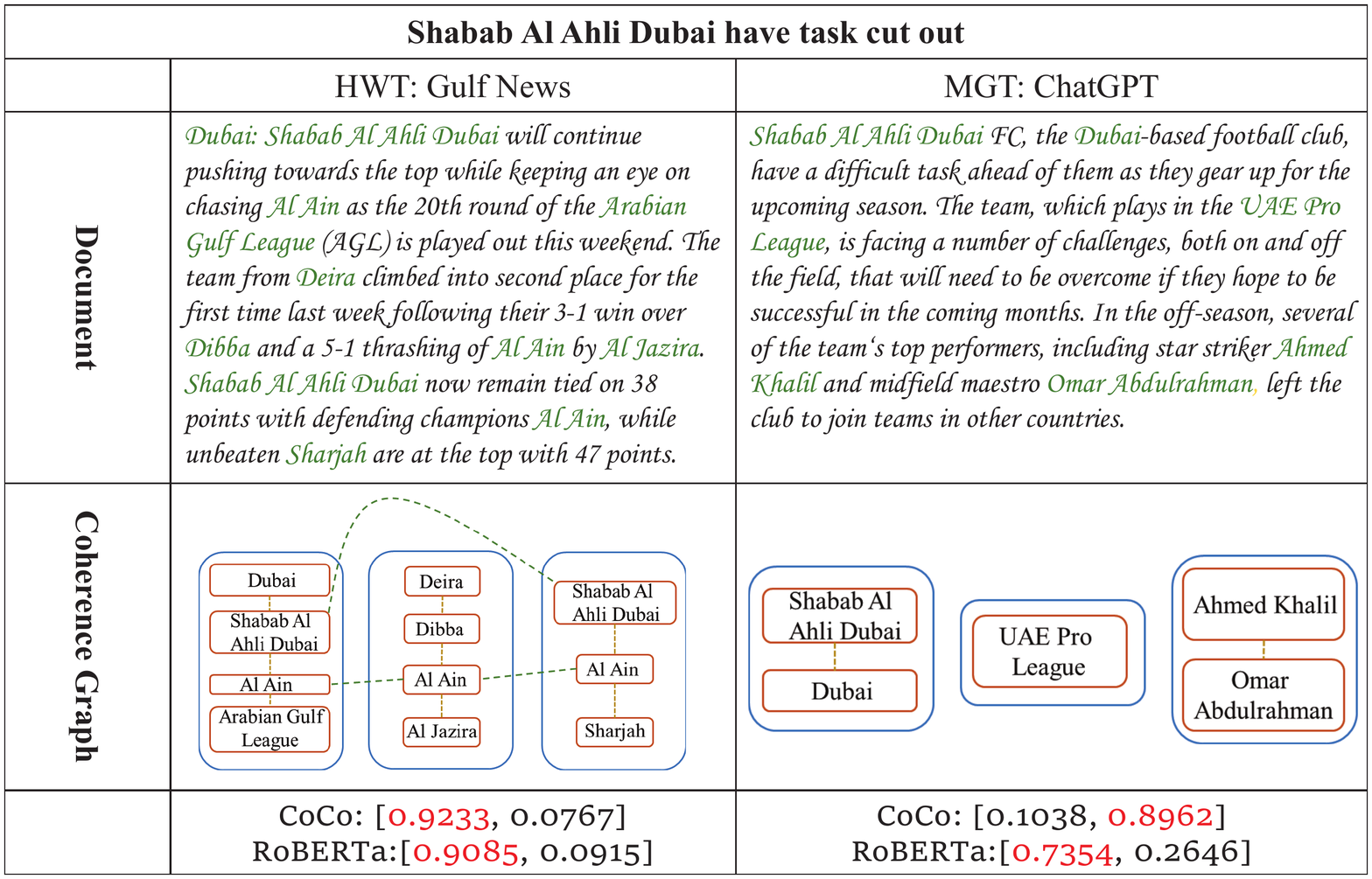}
  \caption{An illustration for case study of our method. Entities in documents are colored green. The blue solid box indicates the sentence. The orange dashed lines are inner edges and green dashed lines are inter edges. Numbers in red indicate the probability of predicted label.} 
  \label{fig:case}
\end{figure}

In this subsection, we conduct a case study with HWT and MGT produced by sensational ChatGPT with the same metadata. 
As illustrated in \figref{fig:case}, we parse two news as coherence graphs.
And we observe that although ChatGPT expresses fluently, it is not coherent from the perspective of coherence graph. 
Hence, \sysname  utilizes the distinctive coherence feature and makes correct predictions.
However, RoBERTa fails to discriminate the MGT without noticing the coherence difference.
This reflects even the most popular and advanced language model could suffer from weak coherence and be detected by \sysname.

\subsection{Token Importance in GPT-3.5 Detection\label{case:token}}

\begin{figure*}[t]
  \centering
  \includegraphics[width=0.83\textwidth]{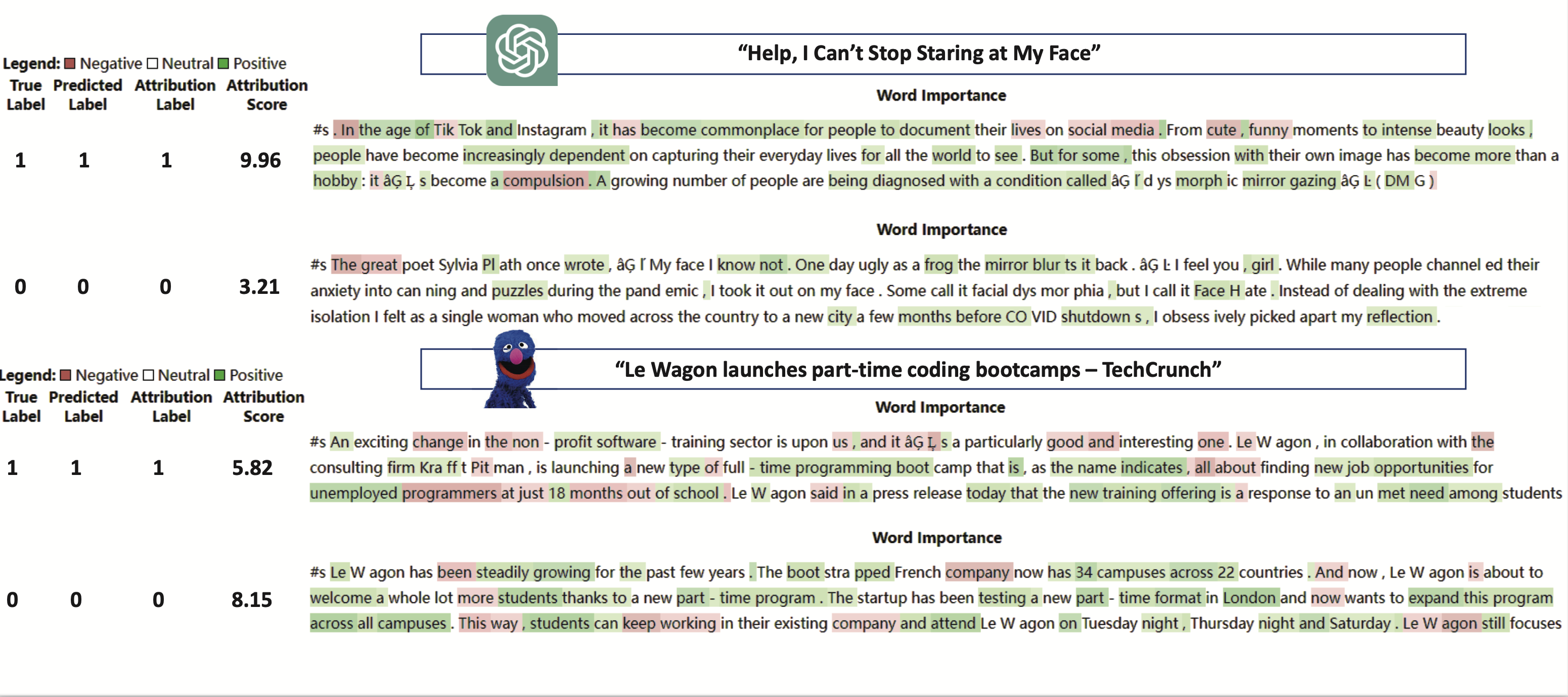}
  \caption{Visualization of token attributions. The first text pair is sampled from GPT-3.5 mixed dataset and the second text pair is from GROVER dataset. The tokens in green represent contributing positively to the predicted label, while those in red contribute negatively. Label "0" represents HWT, and Label "1" represents MGT.}
  \label{fig:explain}
   \vspace{-0.5cm}
\end{figure*}

As shown in Fig. \ref{fig:explain}, we take segments from two text pairs consisting of HWT and its corresponding MGT in GPT-3.5 mixed and GROVER dataset.
It could be noticed that consecutive spans in text generated by GPT-3.5 tend to contribute more to the model decision. 
However, in HWTs, model pays more attention to individual tokens. 
Following this observation, we infer that with the improvement of model scale, LLMs fit extremely well to the corpus so that it generates more general expressions compared with HWTs, which follows certain patterns (always demonstrated by a span of tokens) that could be expected by fine-tuned models.
Thus, barely all the methods show nearly perfect performance on GPT-3.5 dataset.

As for GROVER dataset, more tokens contribute negatively to the model prediction, even if the prediction is correct.
This reflects the deceptive nature of GROVER and explains the reason why it is the hardest dataset in our experiment to some extent.

\section{Static Geometric Analysis on Coherence Graph\label{app:static}}

We have witnessed performance enhancement by applying the graph-based coherence model to the detection model, but how does the coherence graph help detection? 
In this subsection, we apply static geometric features analysis to coherence graph we construct to evaluate the distinguishable difference between HWTs and MGTs with explanation. 
In the following discussion, we take the dataset of GROVER into the analysis.
Some basic metrics of data and the corresponding graph are shown in Table \ref{basic}. 

\begin{table}[!ht]
\centering
\begin{tabular}{l|l|l}
\hline
\textbf{Metric}      & \textbf{HWT} & \textbf{MGT} \\ \hline
Sample Num.          & 4994         & 4991         \\ \hline
Avg. Num. of Token   & 463.2        & 456.0        \\ \hline
Avg. Num. of  Vertex & 43.60        & 32.37        \\ \hline
Avg. Num. of Edge    & 107.4        & 65.44        \\ \hline
\end{tabular}
\caption{Basic metrics of texts and corresponding graphs.}
\label{basic}
\end{table}

Though HWTs and MGTs have approximately the same number of tokens in every text, coherence graph for HWTs has larger scale than MGTs’ with \textbf{34.7\%} more vertexes and \textbf{64.1\%} more edges, which shows that HWTs have more complex semantic relation structures than MGTs.

\subsection{Degree Distribution\label{degreedis}}

Semantically, degree of coherence graph measures the co-occurrence and TF-IDF feature of keywords. 
Moreover, degree distribution shows global coherence because high-degree nodes devote to the main topic and low-degree nodes are the extension.

\begin{table}
    \centering
    \renewcommand\arraystretch{1.3}
    \begin{tabular}{l|c}
    \hline
    \textbf{Metric}        & \textbf{Avg. Degree} \\ \hline
    \textbf{HWT}           & 2.980 \\ \hline
    \textbf{MGT}           & 2.591 \\ \hline
    \end{tabular}
    \captionof{table}{Average of degree (whole dataset). }
    \label{deg}
\end{table}

\begin{figure}[!ht]
\centering
\includegraphics[width=1.0\linewidth]{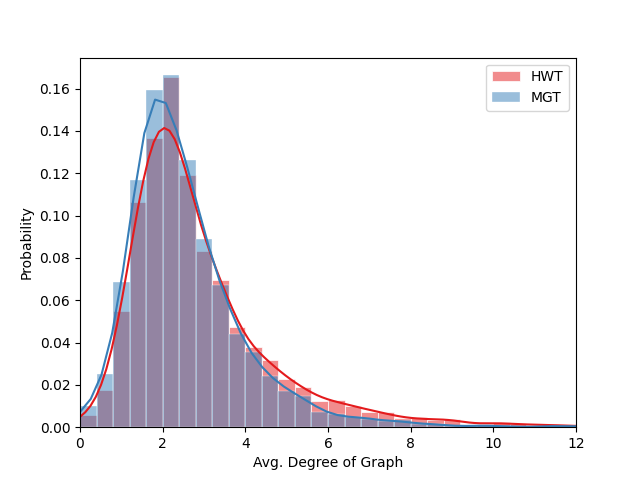}
\caption{Distribution of average degree of graphs.}
\label{fdeg}
\end{figure}

\begin{figure*}[!t]
    \centering
\includegraphics[width=0.8\linewidth]{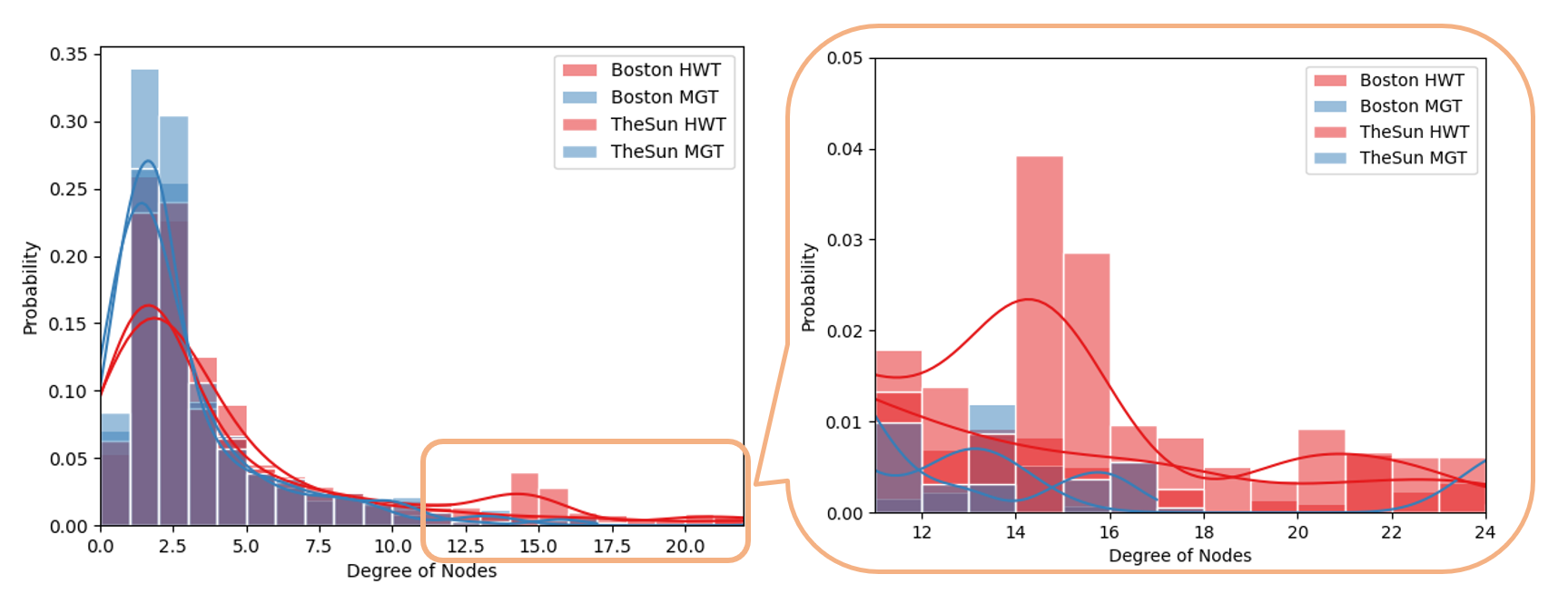}
    \caption{Distribution of degree with different provenance.}
    \label{fig:dpro}
\end{figure*}

As shown in \tabref{deg}, The degree of the graph representation of HWTs is \textbf{2.980}, which is \textbf{15.0\%} larger than MGTs (\textbf{2.591}), which shows disparities of MGTs to form coherent interaction between sentences. 
\figref{fdeg} measures the distribution of each graph's average nodes' degree, showing that the distribution of HWTs has a longer tail than MGTs.

Furthermore, we analyze the distinguishability of degree features when impacted by other factors. One most considerable influences is the style and genre of different provenance. We chose around 60 articles from The Sun\footnote {https://www.thesun.co.uk/} and Boston\footnote {https://www.boston.com/}. Then we use GROVER to mimic their style to generate similar topic news. \figref{fig:dpro} shows the degree distribution of HWTs and MGTs of both provenances.

We use the Jensen–Shannon divergence to evaluate the similarity of the degree distribution. 
The JS-divergence of MGTs mimicking The Sun and Boston is \textbf{0.029}, while the JS-divergence of MGTs and HWTs in Boston is \textbf{0.050}, in The Sun is \textbf{0.061}. The apparent gap shows that degree distribution can robustly detect MGTs and HWTs when impacted by provenance differences.

\subsection{Aggregation}

Aggregation is a shared metric for complex networks and linguistics, depicting how closely the whole is organized around its core. We propose two metrics to evaluate the aggregation of graph-based text representation in our coherence model, the size of the largest connected subgraph and the clustering coefficient. 

\begin{figure}[!ht]
\centering
\includegraphics[width=\linewidth]{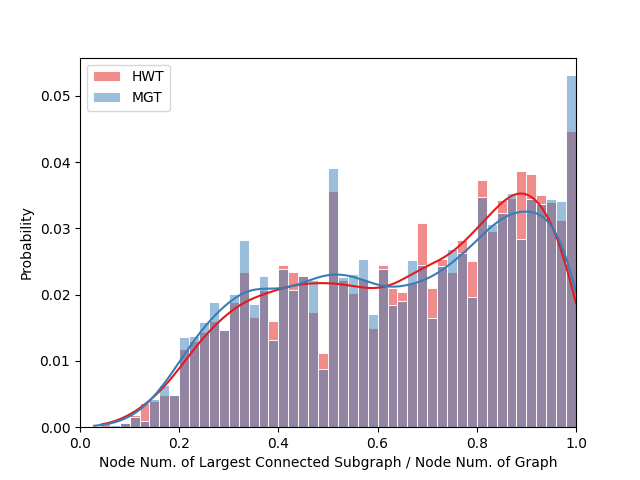}
\caption{Portion of the largest connected subgraph.}
\label{lcs}
\hfill
\centering
\includegraphics[width=\linewidth]{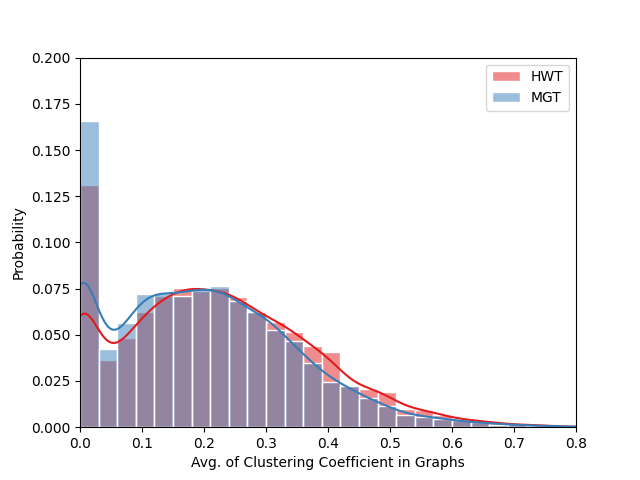}
\caption{Distribution of clustering coefficient.}
\label{fcc}
\end{figure}

In our representation, not all sentences have entities related to others. Hence the graph is an unconnected one. The average number of nodes in subgraphs of MGTs is \textbf{4.49} and of HWTs is \textbf{4.84}. We propose that the size of the largest connected subgraph shows the contents which are closely organized around the topic. Moreover, the size of graphs may be an unfair factor, so we use the portion of nodes in the largest connected subgraph to reflect its size. The average portion in HWTs is \textbf{0.6725} and in MGTs is \textbf{0.6458}. \figref{lcs} shows the distribution of the portion of graphs, and HWTs distribute more high-portion ones than MGTs.

The clustering coefficient represents how nodes tend to cluster. For the entities of texts, clustering evaluates how the author narrates around the central theme. The larger the clustering coefficient is, the tighter the semantic structure is. The average cluster coefficient of the graphs of HWTs is \textbf{0.2213} and of MGTs is \textbf{0.1983}, HWTs is \textbf{11.6\%} better than MGTs. \figref{fcc} shows the distribution.

\subsection{Core \& Degeneracy}

\begin{figure}[!ht]
\centering
\includegraphics[width=\linewidth]{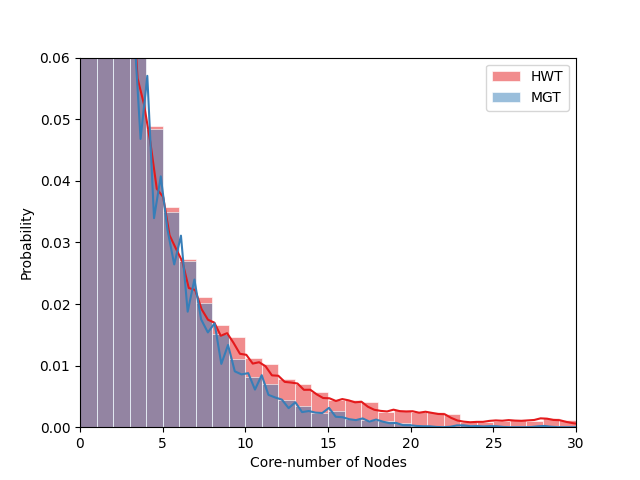}
\caption{Core-number of nodes in graphs}
\label{kc}
\hfill
\centering
\includegraphics[width=\linewidth]{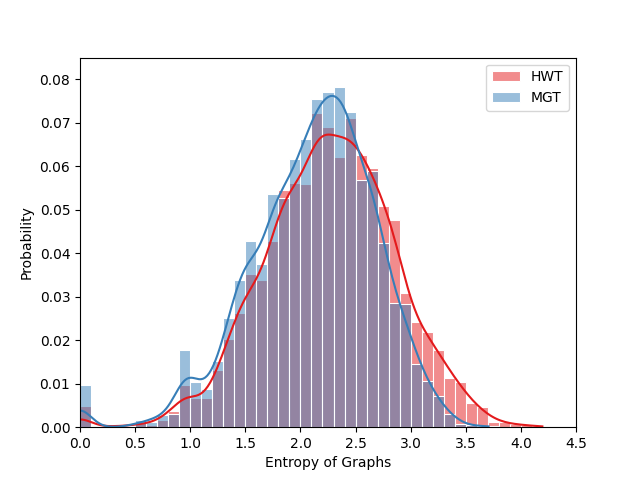}
\caption{Structure entropy of graphs}
\label{fen}
\end{figure}

The degeneracy of a graph is a measure of how sparse it is, and the $k$-core is the subgraph corresponding to its significance in the graph. We propose that,  in our graph representation,  the degeneracy process of graphs equals summarizing texts semantically. The maximum of core-number shows the complexity of hierarchical structure in texts. Furthermore, the distribution of the core-number reflects the overall sparse and is a graph-perspective N-gram module. Based on experiments, the average core-number of HWTs is \textbf{5.772} while MGTs with \textbf{4.458}. HWTs are \textbf{29.5\%} ahead. \figref{kc} is the distribution of the core-number.

\subsection{Entropy}

Entropy is a scientific concept to measure a state of disorder, randomness, or uncertainty. The well-known Shannon entropy is the core of the information theory, measuring the self-information content. For the graph data, network structure entropy defined as the following can examine the information amount of the graph structure.

\begin{small}
\begin{equation}
   Entropy=-\sum_{i=1}^N I_i  \ln I_i = -\sum_{i=1}^N \frac{k_i}{\sum_{j=1}^N k_j} \ln( \frac{k_i}{\sum_{j=1}^N k_j}),
\end{equation}
\end{small}
where $I_i$ is the information content represented by the degree distribution, $N$ is the number of nodes, and $k_i$ is the degree of the $i$-th node.

Global coherence, from our perspective, equals refining more information inside the semantic structure of the whole text, which matches to structure entropy of our graph representation. From our experiments, the structure entropy of HWTs (2.263) is \textbf{6.80\%} larger than MGTs (2.119), which means HWTs obtain more structured information because their semantic information is globally organized. We show the network structure entropy distribution in \figref{fen}.

\section{Exploration on Imbalanced Data}\label{imbalance}

\begin{figure}[!h]
    \centering
    \includegraphics[width=1.0\linewidth]{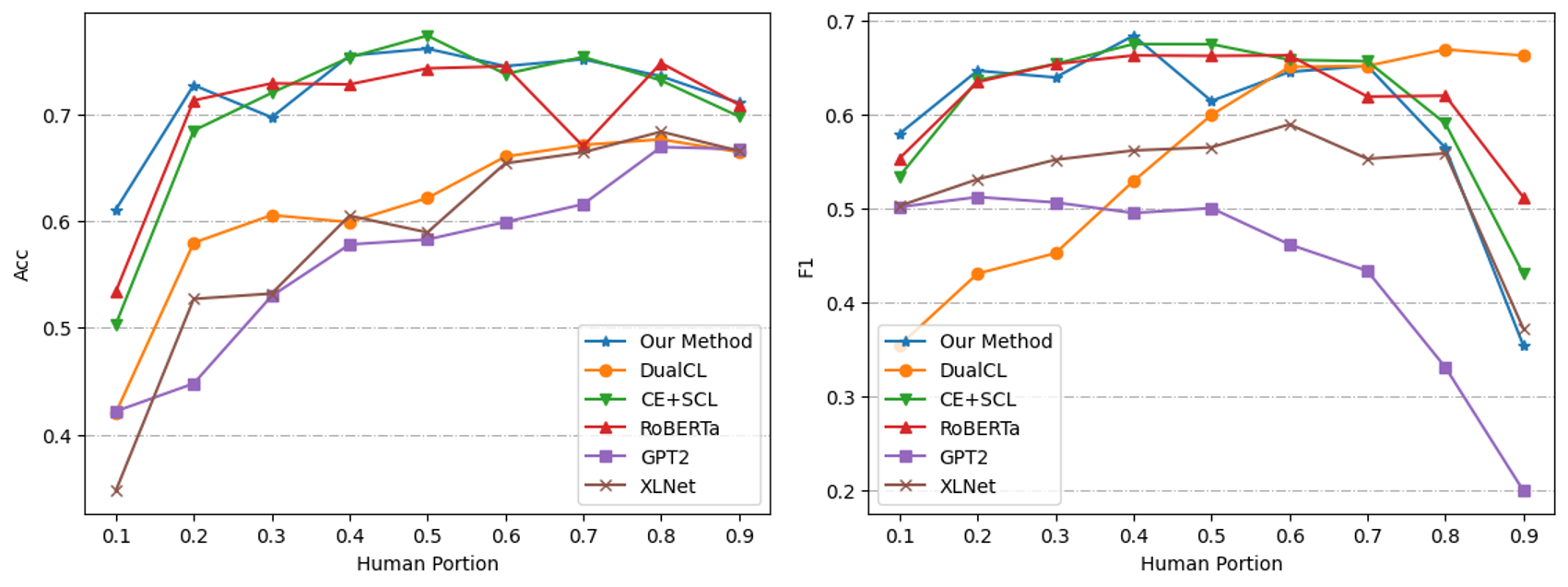}
    \caption{Model comparison results on DL dataset with 9 different human-generated text portions.}
    \label{fig:exp results}
\label{imb}
\end{figure}

Imbalanced distribution in data is another crucial limitation in the task of MGTs detection, which is similar to the low resource limitation. It is imaginable that, with the development of generation technology, MGTs will overwhelmingly dominate low-quality articles since they are easier and faster to generate than human writing. The detection model will face training resources with MGTs as the main part and HWTs as the small part. We test the current models in the imbalanced limitation and find the dramatic decline in accuracy when the ratio of HWTs is less than 30\%, as shown in the \figref{imb}. The test is based on the 10\% GROVER dataset.

All models show poor performance at low HWTs ratios. With a percentage of HWTs of 0.1 (only 100 HWTs in the training set in this case), most of the models have an accuracy below 50\%, which performance is close to random and reflects intolerance for extreme cases. Besides, we find that a high proportion of HWTs also cause a decrease in F1 score to some extent.

\section{Related Work: Graph-based Text Representation}

Graph-of Words (GoW) Model \cite{ turney2002learning, mihalcea2004textrank} is a type graph representation method in which each document is represented by a graph, whose nodes correspond to terms and edges capture co-occurrence relationships between terms. 
Using GoW, keywords can be extracted by retaining the document graph \cite{turney2002learning}. 
Thus, graph representation is sensible to apply in tasks like information retrieval \citep{ Blanco2011GraphbasedTW}, categorization \citep{ malliaros2015graph} and sentiment classification tasks \citep{ huang2019syntax, hou2021graph}. 

Most models enhance classification or detection performance by combining graph representation with neural networks.
Text-GCN \citep{ yao2019graph} first builds a single large graph for the whole corpus, followed by Tensor-GCN \citep{ liu2020tensor} with tensor representation. 
Also, the relation between words varies, and should be treated as different edges. \sysname matches keywords PLM embedding to nodes and sentence representation, considers dealing inner- and inter-sentence relation differently in GCN, and merges the structure graph and flat sequence representation to predict accurately.


\end{document}